\PassOptionsToPackage{dvipsnames}{xcolor} 
\documentclass[journal]{IEEEtran}
\pdfoutput=1 
\ifCLASSINFOpdf
\else
\fi

\usepackage{amsmath}
\usepackage[framemethod=tikz]{mdframed}
\usepackage{lipsum}
\usepackage{cuted}
\usepackage{amssymb}
\usepackage{mathtools}
\usepackage{enumerate}
\usepackage{hhline}
\usepackage{enumitem}
\usepackage{graphicx}
\usepackage{subfigure}
\usepackage{xcolor}
\definecolor{mypink1}{rgb}{0.858, 0.188, 0.478}
\usepackage{todonotes}
\usepackage{algpseudocode}
\usepackage{algorithm}
\usepackage{adjustbox}
\usepackage{color}
\usepackage{soul}

\usepackage{textcomp}

\usepackage[dvipsnames]{xcolor}

\algdef{SE}[SUBALG]{Indent}{EndIndent}{}{\algorithmicend\ }%
\algtext*{Indent}
\algtext*{EndIndent}

\begin{document}

\title{A Robust Asymmetric Kernel Function for Bayesian Optimization, with Application to Image Defect Detection in Manufacturing Systems}

\author{Areej~AlBahar,
      Inyoung~Kim,
      and~Xiaowei~Yue,~\IEEEmembership{Senior Member,~IEEE}
\thanks{This work was partially financially supported by the Department of Defense (DoD)
MEEP program under award N00014-19-1-2728.}
\thanks{A. AlBahar is with the Department of Industrial and Systems Engineering, Virginia Tech, Blacksburg,
VA, 24061, and the Department of Industrial and Management Systems Engineering, Kuwait University, Kuwait. (e-mail: areejaa3@vt.edu).}
\thanks{X. Yue is with the Department of Industrial and Systems Engineering, Virginia Tech, Blacksburg,
VA, 24061. (e-mail: xwy@vt.edu).}
\thanks{I. Kim is with the Department of Statistics, Virginia. Tech, Blacksburg,
VA, 24061. (e-mail: inyoungk@vt.edu).} \thanks{Corresponding author: Xiaowei Yue}
\thanks{\copyright2021 IEEE. Personal use of this material is permitted.  Permission from IEEE must be obtained for all other uses, in any current or future media, including reprinting/republishing this material for advertising or promotional purposes, creating new collective works, for resale or redistribution to servers or lists, or reuse of any copyrighted component of this work in other works. }
}

\markboth{Accepted by IEEE Transactions on Automation Science and Engineering}%
{Shell \MakeLowercase{\textit{et al.}}: Bare Demo of IEEEtran.cls for IEEE Journals}

\maketitle

\begin{abstract}
Some response surface functions in complex engineering systems are usually highly nonlinear, unformed, and expensive-to-evaluate. To tackle this challenge, Bayesian optimization, which conducts sequential design via a posterior distribution over the objective function, is a critical method used to find the global optimum of black-box functions. Kernel functions play an important role in shaping the posterior distribution of the estimated function. The widely used kernel function, e.g., radial basis function (RBF), is very vulnerable and susceptible to outliers; the existence of outliers is causing its Gaussian process surrogate model to be sporadic. In this paper, we propose a robust kernel function, Asymmetric Elastic Net Radial Basis Function (AEN-RBF). Its validity as a kernel function and computational complexity are evaluated. When compared to the baseline RBF kernel, we prove theoretically that AEN-RBF can realize smaller mean squared prediction error under mild conditions. The proposed AEN-RBF kernel function can also realize faster convergence to the global optimum. We also show that the AEN-RBF kernel function is less sensitive to outliers, and hence improves the robustness of the corresponding Bayesian optimization with Gaussian processes. Through extensive evaluations carried out on synthetic and real-world optimization problems, we show that AEN-RBF outperforms existing benchmark kernel functions. 

\end{abstract}

\def\abstractname{Note to Practitioners}
\begin{abstract}
Some industrial systems cannot be accurately represented by physical models. In this situation, data-driven black-box optimization is necessary for advancing the system automation and intelligence. Bayesian optimization is one widely used strategy for learning the global optimum of black-box functions. Bayesian optimization has been applied to robotics, anomaly detection, automatic learning algorithm configuration, reinforcement learning, and deep learning. This paper proposes one new kernel function, named after AEN-RBF. The new kernel function will make Bayesian optimization with Gaussian processes more robust to outliers and lower the data quality barrier of model training. This paper was motivated by the hyperparameters tuning problem of deep learning models for image defect detection in advanced manufacturing, but the method can be easily extended to other applications where kernel functions are needed. Our proposed method is verified by synthetic and real-world optimization problems. 

\end{abstract}

\begin{IEEEkeywords}
Advanced Manufacturing, Bayesian Optimization, Defect Detection, Gaussian Process, Process Optimization.
\end{IEEEkeywords}

\section{Introduction}
\label{sec:intro}

\begin{figure}[!t]   
 \centering
 \includegraphics[width=\columnwidth]{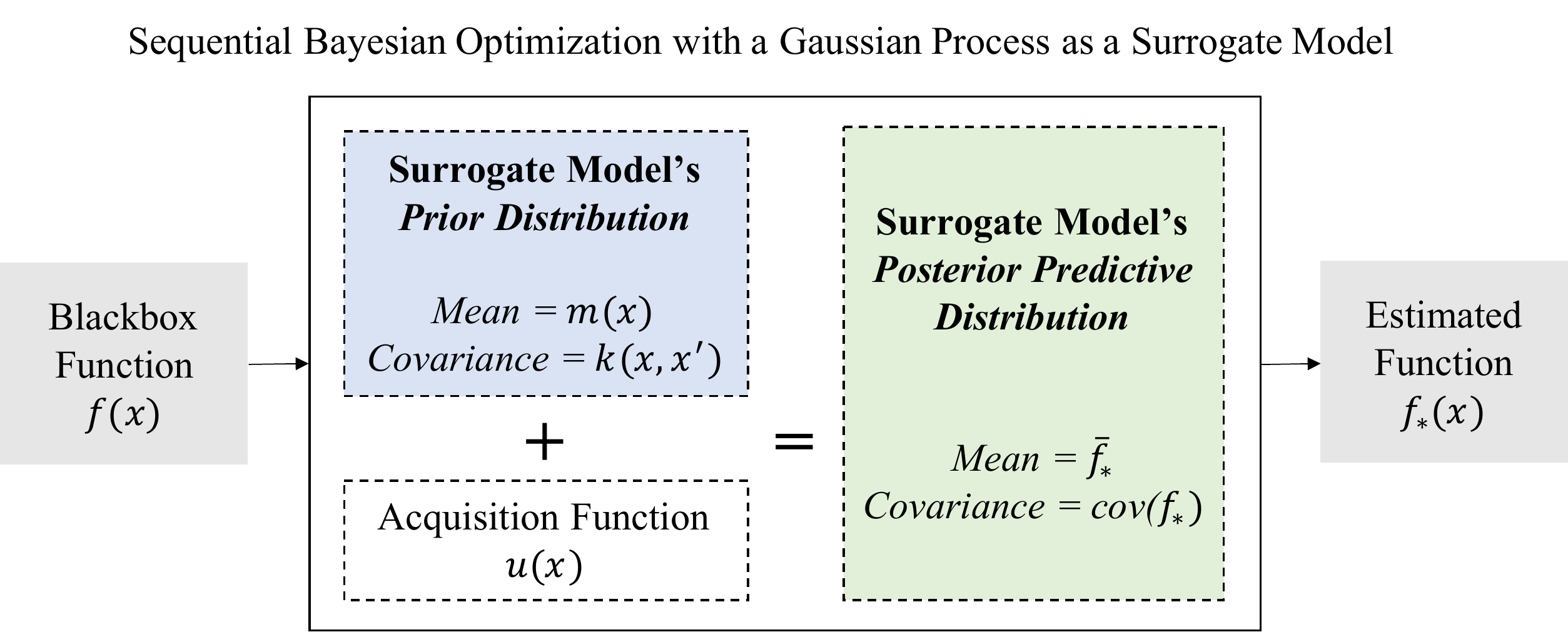}

\caption{\textbf{General approach to Bayesian optimization.}
The unknown black-box function (i.e., objective function) goes through a sequential Bayesian optimization model where it is analyzed and estimated. A surrogate model is built to capture the probabilistic distribution over the objective function (i.e., prior distribution) which is specified by a mean function and a covariance (i.e., kernel) function. An acquisition function is used to determine the next engineering design to update the prior distribution into the posterior distribution. Finally the predictive posterior distribution will be used to calculate the optimum of the objective function.
}
 \label{fig_overview}
\end{figure}

\IEEEPARstart{B}{lack-box} optimization is a class of global optimization where the objective function is expensive to evaluate, has an unknown analytical form, and is in most cases non-convex \cite{pardalos2013handbook}. In many engineering systems, physics-driven models cannot represent the system accurately or even do not exist due to the highly nonlinear complex structures and uncertainty of system knowledge. To optimize these systems, data-driven black-box optimization can estimate the objective function and find the global optimum in a minimum number of function evaluations.
Bayesian optimization (BO) is a sequential model-based optimization algorithm used to find the global minimum of a black-box function~\cite{mockus2012bayesian}. As shown in Fig.~\ref{fig_overview}, Bayesian optimization models are defined by a surrogate model and an acquisition function.
Surrogate models are probabilistic models that define and estimate unknown functions. The most widely used surrogate model is the Gaussian process (GP). Gaussian processes are used to define probability distributions over functions where each distribution is specified by a mean function $m(x)$ and a positive semi-definite kernel function $k(x,x^{\prime})$~\cite{williams2006gaussian}. Bayes' rule is used to derive the posterior distribution of the GP given the prior and the observations. 
Acquisition functions are mathematical equations used to trade off exploration and exploitation to iteratively determine the next-step query point over a bounded domain $x \in \cal{D}_X$ \cite{7352306}. The acquisition functions are closely associated with sequential optimal designs in statistics \cite{sant2018} or active learning functions in machine learning \cite{yue2020active}. More discussion related to surrogate models and acquisition functions can be found in Section \ref{sec:related}.

Bayesian optimization has been extensively used in a wide range of areas including integrated system design \cite{torun2018global}, industrial control systems \cite{neumann2019data}, robotics \cite{kim2020proactive} and autonomous systems \cite{martinez2018funneled}. In particular, Bayesian optimization is super advantageous for hyperparameter tuning and optimization of automated machine learning models \cite{feurer2019hyperparameter,bergstra2011algorithms}. 
 Hyperparameters are parameters that are not automatically optimized or directly derived during training machine learning models. These hyperparameters are usually manually adjusted to realize the optimal performance of machine learning models, which has been very efficient in regimes where only limited number of trials are possible. Popular hyperparameter tuning methods include grid search, random search, and Bayesian optimization \cite{feurer2019hyperparameter, Wu2019HyperparameterOF}. Grid search is a method that tests and evaluates every single combination of hyperparameters, although reaching the best combination of the hyperparameters is assured, it is very time consuming and not efficient for high-dimensional hyperparameters. Random search \cite{bergstra2011algorithms} on the other hand is a method that selects and tries random combinations of hyperparameters. Random search is computationally inexpensive, however finding the best combination of hyperparameters is not guaranteed. Bayesian optimization is a probabilistic approach that implements Bayes' theorem to find and select the optimal combination of hyperparameters in a minimum number of function evaluations. 
More recently, Bayesian hyperparameter optimization has been used in optimizing different engineering processes. For instance, transfer Bayesian optimization has been applied in optimizing design parameters of manufacturing processes~\cite{BO_Design}. In production processes on the other hand, Bayesian optimization has been applied in optimizing the multi-component predictive systems tasks~\cite{Chemical}.

In the presence of outliers (i.e., unusual data located far from the mean value), Bayesian optimization methods tend to be inefficient in optimizing the black-box function and finding its global optimum. In particular, Gaussian processes are non-robust probabilistic models and are known to be prone to noise (i.e. outliers). Furthermore, outliers are ubiquitous or even inevitable in advanced manufacturing systems. For example, actuator uncertainty, part uncertainty, modeling uncertainty, and unquantified errors co-exist in the composite aircraft assembly process \cite{yue2018surrogate}; signal-dependent noise and spikes are embedded with the Raman spectroscopy and other photon-generated spectroscopy in nanomanufacturing processes \cite{yue2016generalized,yue2017wavelet}. In these systems, Bayesian inference with Gaussian processes produces sporadic predictive models that inaccurately infer the global optimum of the objective function. Therefore, building robust surrogate models for Bayesian optimization is of great importance. With the critical effect of covariance functions on the performance of surrogate models for Bayesian optimization, the robustness of such functions is strongly needed.

Kernels, also known as covariance functions, are functions that measure the similarity between two data points over a given space. There are two types of kernels, stationary and non-stationary \cite{rasmussen2003gaussian}. Stationary kernels are functions of distances (i.e. invariant to translations), examples of stationary kernels include Radial Basis Function (RBF) and Matern. Examples of non-stationary kernels include Linear and Polynomial kernels. 

Non-stationary kernel functions allow Gaussian process models to adapt to variability changes with location and time domains. However, they have limitations such as occasional creation of nearly singular covariance matrices and inflexibility in modelling high-order input interactions as in the case with Polynomial kernels ~\cite{fang2016flexible}, and complex numerical Monte-Carlo integration as in the case with convolutional and spectral kernel functions~\cite{conv_kernel, spectral_kernel}. Stationary kernel functions on the other hand, are highly differentiable and simpler to evaluate than non-stationary kernels. Kernel functions have a significant impact on the accuracy and shape of the GP surrogate models. In this paper, we focus on stationary kernel functions and tackle their challenges and limitations.

Radial Basis Functions (RBFs) are kernel functions that are widely used in optimizing black-box functions~\cite{williams2006gaussian,Gutmann2001ARB}. Despite being infinitely differentiable, RBF kernels are very smooth and tend to be very sensitive to outliers thus constructing sporadic models. Sporadic models produce inaccurate estimated functions, which will then create imprecise and unreliable predictive models. Although Matern kernel function solves the smoothness problem of the RBF kernel, its sensitivity to outliers is still a major challenge. Therefore, developing a robust kernel function is critical to the accuracy and performance of the Bayesian optimization with GP surrogates.

In the area of robust kernel functions, there exists some research concerning asymmetrical kernels in other domains such as~\cite{Mackenzie} which introduces the use of a Gamma density function instead of the Gaussian (i.e., RBF) kernel function to improve boundary errors. Another research concerning the use of asymmetric kernels in Gaussian Processes is the use of multiple Gaussian kernel functions with center averages as inputs instead of single data points and a different lengthscale hyperparameter per kernel~\cite{pintea2018asymmetric}. However, such kernel functions are tailored for specific applications and are not suited for Bayesian optimization and a major limitation of stationary kernel functions used in Bayesian optimization when outliers are present still exists.

In this paper, we propose a novel Asymmetric Elastic Net Radial Basis Function (AEN-RBF) kernel function to produce robust Gaussian process surrogate models for Bayesian optimization. We introduce two parameters, the weight parameter $\lambda$ and the skewness parameter $\alpha$ to improve the similarity measure in kernel definition. We propose a skewed weighted sum of squared Euclidean and Manhattan distances to reduce the GP surrogate model’s sensitivity to outliers and thus enhance its robustness.

Our contributions are summarized as follows.
\begin{itemize}
    \item[1.] We propose AEN-RBF, an Asymmetric Elastic Net Radial Basis Function and prove its validity as a positive semi-definite kernel function.
    \item[2.] We provide theoretical investigations that show the convergence and computational properties of the Bayesian optimization model when AEN-RBF kernel function is used. We also prove that AEN-RBF kernel function provides a lower mean prediction error in comparison with the RBF.
    \item[3.] The proposed AEN-RBF kernel is evaluated in optimizing four synthetic functions and in hyperparameter tuning of deep learning models on real-world datasets. We show that AEN-RBF outperforms benchmark kernel functions. 
\end{itemize}

The remainder of this paper is organized as follows: Section \ref{sec:related} illustrates the background of Bayesian optimization from three perspectives (e.g., surrogate models, kernels, and acquisition functions). Section \ref{sec:overview} describes the proposed AEN-RBF kernel and computational algorithm for parameter estimation, as well as proves the theoretical comparison between AEN-RBF and RBF kernels. Section \ref{sec:property} discusses the properties of the AEN-RBF kernel function, such as convergence, computational complexity, and extendability to other applications. Section \ref{sec:results} conducts the numerical study and compares the proposed AEN-RBF kernel against the benchmark kernels. Section \ref{sec:case_study} presents the case study of image defect detection in advance manufacturing. Finally, a brief summary is provided in Section \ref{sec:conclusions}.

\section{Background and related work}
\label{sec:related}

Bayesian optimization, a statistical model based on Bayes rule \cite{Bayes:63}, can find the global optimum of a black-box function while using a minimum number of function evaluations. The process of Bayesian optimization starts by building a probabilistic data-driven approximation of the objective function called the \textit{surrogate model}. It then uses an \textit{acquisition function} to search for the next best point to be evaluated and update the \textit{surrogate model}. The acquisition function is cheap to evaluate and trades off exploitation (i.e., sampling at points where the surrogate model predicts a high value of the objective function) and exploration (i.e., sampling at points where the surrogate model predicts a high value of the model's uncertainty). In the following subsections, we summarize the important components of Bayesian optimization.

\subsection{Surrogate models}
Surrogate models, also called metamodels or emulators, are data-driven approximations of the objective function. Surrogate models are used to model prior beliefs about the objective function which are then updated into posterior predictive distributions. One surrogate model used in Bayesian optimization is random forests~\cite{7352306}. 
Random forests are ensembles of decision tree predictors where each tree represents a randomly sampled vector of the data.
Another example of a Bayesian optimization surrogate model is to integrate the tree-structured Parzen estimator (TPE) with the Gaussian process ~\cite{bergstra2011algorithms}, where it is used to generate surrogate models of the objective function via Bayesian reasoning. Other surrogate models include physics based surrogates \cite{loose2009surrogate}, hierarchical linear model \cite{wu2016bayesian}, etc. 
The most widely used surrogate model and the one that uses Bayes rule is the Gaussian process (GP)~\cite{williams2006gaussian,rasmussen2003gaussian,pourhabib2014bayesian,bergstra2011algorithms,sant2018}, also called Kriging \cite{sant2018,yue2018surrogatecontrol}. GP surrogate models are random probabilistic processes for which the random variables are modelled as a multivariate normal distribution. 
A GP starts with a prior distribution that is defined by a mean function and a kernel (i.e., a covariance function). The prior distribution is then updated into the posterior distribution by the end of the optimization process. In this paper, we focus on the Gaussian process as the surrogate model due to its flexibility, tractability, and nice theoretical properties \cite{williams2006gaussian}. 

\subsection{Kernels of Gaussian processes}
Kernels (i.e., covariance functions) are critical components of the GP surrogates in Bayesian optimization. Stationary kernels are distance based functions that calculate the difference between two kernel inputs. A radial basis function (RBF) ~\cite{Gutmann2001ARB}, is one of the most commonly used kernels in Bayesian optimization and function estimation. RBF is a function of a squared Euclidean distance and is parameterized by a lengthscale that governs its smoothness. Another well known stationary kernel function is Matern. Matern is a generalization of RBF with an additional smoothness parameter. Matern is equivalent to 
RBF when that extra smoothness parameter goes to infinity \cite{rasmussen2003gaussian, 7352306}.  Common non-stationary kernels include but are not limited to Linear and Polynomial kernel functions. In this paper, we focus on developing one new kernel to make the surrogate model more robust to outliers and obtain the resilient Bayesian optimization. More specifically, we focus on RBFs since they are widely used and have few hyperparameters to tune as opposed to Matern kernel functions.

\subsection{Acquisition functions}
Acquisition functions are inexpensive to evaluate and are used to find the next best point to be evaluated. Acquisition functions are used to update the prior distribution of the Gaussian process model into the posterior distribution. One acquisition function is the probability of improvement (PI)~\cite{wilson2018maximizing}.
PI works by finding the probability that a point will lead to the improvement of the objective function. 
The expected improvement (EI) is one of the most widely used acquisition functions \cite{Donald}. The main idea is to identify the input-output relationships, and subsequently select design points to maximize the expected improvement on the objective function.
The upper confidence bound (UCB) can trade off exploitation and exploration through cumulative regret bounds to improve acquisition performance. \cite{Auer}. It is worth noting that the acquisition function in Bayesian optimization shares the same idea with active learning and sequential optimal design \cite{yue2020active}. In machine learning domain, active learning iteratively selects the next data point for maximizing information acquisition to improve model performance. Sequential optimal design, which is the term in statistics domain, can explore the most informative new experimental samples according to the current existing data. More detailed literature review on active learning and sequential design refers to \cite{yue2020active}.

\section{Asymmetric Elastic Net Radial Basis Function}
\label{sec:overview}

In this section, we propose the new kernel, Asymmetric Elastic Net Radial Basis Function (AEN-RBF), and derive the computational algorithms for hyperparameter learning. We also prove that the proposed AEN-RBF kernel can realize smaller mean squared prediction error than the RBF kernel under mild conditions.

\subsection{Bayesian Optimization with RBF Gaussian Processes}
A sequential model-based optimization (SMBO) algorithm is used to minimize unknown and expensive to evaluate functions (i.e., black-box functions). A typical Bayesian optimization algorithm with a Gaussian process surrogate model is shown in Fig.~\ref{fig_overview}. It consists of two components. The first component is the prior distribution over the objective function, which is defined by a mean function and a kernel function. The second component is the acquisition function which is used to select the next point to be evaluated and update the GP's prior distribution into the posterior.

Let $f(\boldsymbol{x})$, $m(\boldsymbol{x})$, and $k(\boldsymbol {x},\boldsymbol{x}^{\prime})$ be the black-box function to be optimized, the mean function, and the kernel function, respectively. We model $f(\boldsymbol{x})$ as a Gaussian process as shown in Equation~\ref{eq:GP}, with the mean function in Equation~\ref{eq:mean} and the kernel function in Equation~\ref{eq:kernel}.
\begin{gather}
    \label{eq:GP}
    f(\boldsymbol{x})\sim GP(m(\boldsymbol{x}) ,k(\boldsymbol{x},\boldsymbol{x}^{\prime})),\\
    \label{eq:mean}
    m(\boldsymbol{x}) = \mathrm{E}\{f(\boldsymbol{x})\},\\
    \label{eq:kernel}
    k(\boldsymbol{x},\boldsymbol{x}^{\prime}) = \mathrm{E}[\{f(\boldsymbol{x}) - m(\boldsymbol {x})\}\{f(\boldsymbol{x}^{\prime})-m(\boldsymbol{x}^{\prime})\}].
\end{gather}

The ultimate goal of Bayesian optimization is to find ${\boldsymbol x}^*$ such that $f(\boldsymbol{x})$ is minimized over its domain ($\cal{D}_X$),
\begin{equation*}
\boldsymbol{x}^* = argmin f(\boldsymbol{x}) \:\:\: \text {for}\: \boldsymbol{x} \in \cal {D}_X.
\end{equation*} 
The expected improvement acquisition function, $u(\boldsymbol{x})$, is defined as $u(\boldsymbol{x}) = \mathrm{E} [\max (0, f^* - f(\boldsymbol{x})]$, where $f^*$ is the optimal function value so far. The expected improvement works by selecting the next best point to be evaluated $\boldsymbol{x}_{next} = argmax \: u(\boldsymbol{x})$ for $\boldsymbol{x} \in {\cal D}_X$. The expected improvement is used to update the prior distribution of the GP model into the posterior distribution and hence produce a predictive model of the objective function. The posterior predictive model is then used to obtain the global optimum of the objective function.

The Radial Basis Function (RBF), also known as the \textit{squared exponential} kernel function, is the most widely used covariance function in Gaussian process regression and machine learning~\cite{williams2006gaussian}. Equation~\ref{eq:RBF_kernel} represents the mathematical formula of the RBF, where $D_{E}$ is the squared Euclidean distance shown in Equation~\ref{eq:EucDis}. The hyperparameters \textit{l} and $\sigma_{0}^2$ represent the lengthscale and the variance, respectively. Hyperparameters are parameters of the mean and kernel functions and are optimized using maximum likelihood estimate (MLE) approach.

\begin{equation}
\label{eq:RBF_kernel}
k_{RBF}(\boldsymbol{x}, \: \boldsymbol{x}^{\prime}) = \sigma^2_{0} \times e^{-\frac{D_{E}(\boldsymbol{x},\boldsymbol{x}')}{2l^2}} 
\end{equation} 

\begin{equation}
\label{eq:EucDis}
D_{E}(\boldsymbol{x},\boldsymbol{x}') = {\Bigg(\sqrt{\sum_{j=1}^{p} {({x}_{ja} - x_{jb}^{\prime})}^2}\:\: \Bigg)}^2
\end{equation} 

The RBF is an infinitely differentiable kernel function. However, a major drawback of using RBF as a kernel function is that it is very smooth and sensitive to outliers. The sensitivity to outliers produces irregular and sporadic surrogate models that inaccurately infer a function. Therefore, we propose the new AEN-RBF kernel function.

\subsection{The Asymmetric Elastic Net Radial Basis Function (AEN-RBF) Kernel Function}

In order to have a surrogate model for Bayesian optimization that is robust and less sensitive to outliers, we investigate a couple of state-of-the-art regression models. The sparsity encouragement in regression may help us formulate the idea and propose a new kernel to enhance the robustness. More specifically, we look at \textit{Lasso} \cite{tibshirani1996regression} and \textit{Elastic Net} \cite{Zou05regularizationand}. 
Lasso regression model uses L1 regularization method. The L1 regularization method adds an absolute value of the magnitude of the regression coefficients to the cost function as a penalty. The addition of this penalty is found to make the model less sensitive to outliers, and therefore generates sparse regression models~\cite{tibshirani1996regression}. On the contrary, L2 regularization, also known as \textit{Ridge} regularization, can better handle the correlation between different variables especially when the number of variables is not very large~\cite{ridge}. L2 regularization is also found to be more stable numerically than its counterpart, the L1 regularization.
The Elastic Net regression model uses a weighted sum of L1 and L2 regularization methods to take the advantages of both regularization methods~\cite{Zou05regularizationand}. 

Inspired by the idea of Elastic Net, we introduce \textit{AEN-RBF}, an Asymmetric Elastic Net Radial Basis Function. \textit{AEN-RBF} is a skewed kernel function with a weighted summation of squared Manhattan and Euclidean distances. The Manhattan distance, also known as the absolute value distance, results in a robust kernel function and a sparser surrogate model especially when dealing with high dimensional data. Equation~\ref{eq:ManDis} represents the mathematical formula of the squared Manhattan distance $D_{M}$. Equation~\ref{eq:ENDis} represents the elastic net distance $D$, which is the weighted summation of squared Manhattan distance $D_{M}$ and squared Euclidean distance $D_{E}$.

\begin{equation}
\label{eq:ManDis}
D_{M}(\boldsymbol{x},\boldsymbol{x}') = {\Big(\sum_{j=1}^{p}{\lvert {x}_{ja} - x_{jb}^{\prime}\rvert} \: \Big)}^2 
\end{equation} 

\begin{equation}
\label{eq:ENDis}
D(\boldsymbol{x},\boldsymbol{x}') = \lambda \cdot D_{M}(\boldsymbol{x},\boldsymbol{x}') + (1-\lambda) \cdot D_{E}(\boldsymbol{x},\boldsymbol{x}')
\end{equation}

The elastic net distance can realize the trade off between Manhattan distance and the Euclidean distance, which corresponds to exploitation and exploration, respectively. We make full use of it when we propose the AEN-RBF kernel function.
Equation~\ref{eq:AEN_RBF_kernel} represents the mathematical formula of AEN-RBF kernel function. 

\begin{equation}
\label{eq:AEN_RBF_kernel}
k_{AEN{\text -}RBF}(\boldsymbol{x}, \boldsymbol{x}^{\prime}) = \sigma^2_{0} \times
\begin{cases}
  e^{-(1-\alpha) \cdot \frac{ D(\boldsymbol{x}, \boldsymbol{x}^{\prime})}{l^2}} & \text{for}\: x_{ja} < x_{jb}^{\prime} \\
  e^{-\alpha \cdot \frac{D(\boldsymbol{x}, \boldsymbol{x}^{\prime})}{l^2}} & \text{for}\: x_{ja} \geq x_{jb}^{\prime}
\end{cases}
\end{equation}

In order to improve the robustness of the RBF kernel function to outliers, we introduce two new parameters, the weight parameter, \textit{lambda} $(\lambda)$, and the skewness parameter, \textit{alpha} $(\alpha)$. The weight parameter, $\lambda$, is between 0 and 1 and is used to scale squared Euclidean and Manhattan distances. The skewness parameter, $\alpha$, is a measure of the asymmetry of the kernel function and is also between 0 and 1 in such a way that when $\alpha = 0.5$ the resulting kernel function is symmetric, whereas when $\alpha > 0.5$ or when $\alpha < 0.5$ the resulting kernel function is skewed to the left and right, respectively.

\begin{figure}[!t]   
 \centering
 \includegraphics[width=\columnwidth]{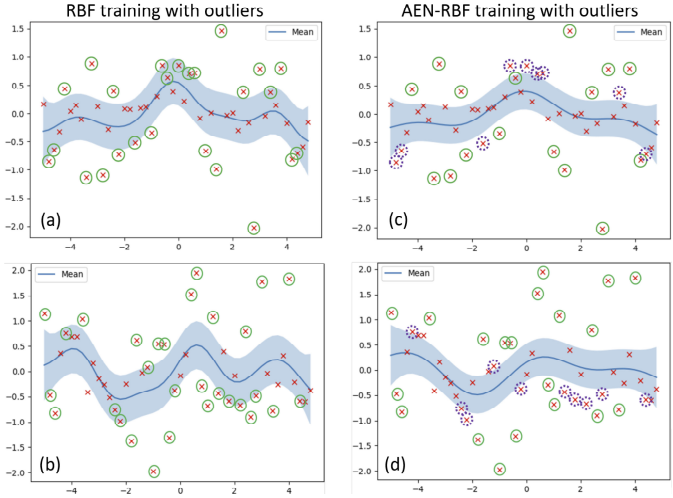}
 \caption{\textbf{Gaussian process surrogate models in the presence of outliers.} (a-b) represent Gaussian process surrogates when RBF kernel function is used, while (c-d) represent Gaussian process surrogates when AEN-RBF kernel function is used. (a,c) belong to outlier scenario one and (b,d) belong to outlier scenario two. The green circles represent the induced outliers, whereas the purple dashed circles represent the excluded outliers in the posterior distribution of the GP when AEN-RBF kernel function is used.
}
 \label{fig:noise}
\end{figure}

Fig.~\ref{fig:noise} represents the posterior distribution of the Gaussian process regression model when RBF and AEN-RBF kernel functions are used. From Fig.~\ref{fig:noise}, we demonstrate the robustness of our AEN-RBF kernel function in comparison with the RBF when using the two experiment scenarios. We can find that AEN-RBF kernel function can reduce the impact of outliers on the posterior distribution of the GP. More specifically, when using RBF as a kernel function, the confidence bounds are enlarged due to the presence of outliers, whereas when AEN-RBF kernel function is used, the confidence bounds are not sensitive to outliers resulting in a number of outliers excluded from the posterior distribution, thus yielding smaller confidence bounds. The excluded outliers are shown in purple dashed circles. These figures show how flexible and robust our AEN-RBF kernel in approximating the objective function when outliers exist.

\textbf{Proposition 1.} The AEN-RBF kernel denoted as $k_{AEN-RBF}$ is a valid kernel function and its corresponding \textit{Gram matrix}, $K_{AEN-RBF}$, is positive semi-definite. \\
\textbf{\textit{Proof.}} Let $k_{D_E}$ and $k_{D_M}$ be the RBF and the squared Manhattan distance based kernel functions, respectively.
We decompose the AEN-RBF kernel function as follows:

\begin{equation*}
\begin{split}
k_{AEN-RBF}(x,x')=\sigma^2_{0} \times e^{[-(1-\alpha)(\lambda D_M +(1-\lambda)D_E)] {\bf 1}(x<x')} \\ \times e^{[-\alpha(\lambda D_M +(1-\lambda)D_E)] {\bf 1}(x\ge x')}.
\end{split}
\end{equation*}
Since the indicator function is a constant kernel function and the linear combination and multiplication of two kernel functions are also a kernel function, then the AEN-RBF kernel is a valid kernel function. It then follows that the \textit{Gram matrix} $K_{AEN-RBF}$ is positive semi-definite by the property of kernel functions.

\subsection{Computational Algorithm for Parameter Estimation}
In our AEN-RBF kernel function, the hyperparameters are estimated using the maximum likelihood estimation approach (MLE)~\cite{williams2006gaussian}. MLE is a special case of the maximum a \textit{posteriori} (MAP) estimate approach. By denoting $\theta$ as the set of hyperparameters to be optimized, we can express the log marginal likelihood of the GP model as $\log p(\boldsymbol{y}|X,\theta)$. We estimate $\theta$ by maximizing the log marginal likelihood (i.e., using its gradient with respect to the hyperparameters $\theta$) of the GP model as follows: 
\begin{gather*}
\frac{\partial }{\partial \theta} \log p(\boldsymbol{y}|X,\theta)\\=\frac{1}{2}tr[\{ (K^{-1}(\boldsymbol{y}-m(\boldsymbol{x}))) (K^{-1} (\boldsymbol{y}-m(\boldsymbol{x})))^T -K^{-1}\} \frac{\partial K} {\partial \theta}],
\end{gather*}
where $\theta$ represents $\sigma^2_0$, $l$, $\lambda$, $\alpha$, and $\sigma_{\epsilon}^2$ hyperparameters.

In the case where the mean function $m(\boldsymbol{x})=\boldsymbol{0}$, the predictive function value ${f}_*$ corresponding to the test input $x_*$ can be sampled from the normal probability distribution $N(\bar{f}_*,cov(f_*))$, where $\bar{f}_*$ and $cov(f_*)$ are the posterior predictive mean and covariance, respectively. The mathematical formulations of $\bar{f}_*$ and $cov(f_*)$ are shown in Equation~\ref{eq:pred_mean_cov}.
\begin{multline}
\label{eq:pred_mean_cov}
\bar{f}_* =k(\boldsymbol{x}_*,\boldsymbol{x})[K+\sigma_{\epsilon}^2 \textbf{I}_n]^{-1}\boldsymbol{y}, \\
cov(f_*)=k(\boldsymbol{x}_*,\boldsymbol{x}_*)-k(\boldsymbol{x}_*,\boldsymbol{x})[K + \sigma_{\epsilon}^2 \textbf{I}_n]^{-1} k(\boldsymbol{x},\boldsymbol{x}_*).
\end{multline}

After finding the optimal set of hyperparameters, $\theta$, and the respective posterior predictive mean and covariance of the GP over the objective function, the estimated global minimum value $\boldsymbol{x}^*$ can be found and its corresponding $y(\boldsymbol{x}^*)$. The detailed Bayesian optimization procedure with AEN-RBF Gaussian process kernel function is shown in Algorithm~\ref{BO_AEN-RBF}.

The AEN-RBF kernel function has four hyperparameters to tune namely the lengthscale parameter denoted as \textit{l}, the variance parameter denoted as $\sigma_{0}^2$, the weight parameter denoted as $\lambda$, and the skewness parameter denoted as $\alpha$.
The AEN-RBF kernel function's hyperparameters are optimized during Bayesian optimization in order to get the posterior predictive distribution of the GP model. The hyperparameters are optimized using the maximum likelihood estimation approach (MLE). According to prior work, we use $\lambda$ values ranging between 0.1 and 0.9 and $\alpha$ values ranging between 0.25 and 0.75.

\begin{algorithm}

  \caption{Bayesian Optimization with AEN-RBF Gaussian Process Kernel Function}
  \label{BO_AEN-RBF}
  \textbf{Input:} A vector of z initial points $X_z = [\boldsymbol{x}_1, .., \boldsymbol{x}_z]$, the prior mean $m(\boldsymbol{x})$ and kernel $k(\boldsymbol{x}, \boldsymbol{x'})$ functions of the GP model, the acquisition function $u(\boldsymbol{x})$, and the maximum number of iterations ($n$).
  \begin{algorithmic}[1]
      \State Find the initial function values $f(\boldsymbol{x}_1), .., f(\boldsymbol{x}_z)$
      \State Estimate the hyperparameters set $\theta$ by maximizing the log marginal likelihood
      \State Update the mean and covariance of the GP model using Equation~\ref{eq:pred_mean_cov}
      \For{\texttt{$i = z+1, .., n$}}
        \State Select the next point to be evaluated, $\boldsymbol{x}_i$, such that
        \Indent $\boldsymbol{x}_i = argmax \: u(\boldsymbol{x})$ for $\boldsymbol{x} \in \cal {D}_X$
        \EndIndent
        \State Evaluate $f(\boldsymbol{x}_i)$ at $\boldsymbol{x}_i$ 
        \State Update the hyperparameters set $\theta$ by maximizing \Indent the log marginal likelihood
        \EndIndent
        \State Find ${y_i} = f(\boldsymbol{x}_i) + \epsilon_i$
        \State Update the mean and covariance of the GP model 
        \Indent using Equation~\ref{eq:pred_mean_cov}
        \EndIndent
        \State Update the so far optimal input point $\boldsymbol{x}^*$ and the 
        \Indent corresponding $y^*$
        \EndIndent
      \EndFor
      \State \textbf{Return} the Gaussian process posterior predictive mean and covariance, and the global optimum $y^*$ at $\boldsymbol{x}^*$ 
  \end{algorithmic}
\end{algorithm}

\subsection{Theoretical Comparison Between AEN-RBF and RBF Kernels}

\textbf{Proposition 2.} Let $\boldsymbol{o}$ and $\boldsymbol{o'}$ be the $p$ dimensional random noise variables following $p_{\boldsymbol{o}}(\boldsymbol{o})$ and $p_{\boldsymbol{o'}}(\boldsymbol{o'})$, respectively, and let $p_{\boldsymbol{x}}$ be the probability density function of $\boldsymbol{x}$. We can then show the following:\\ 
(\textit{i}) if $p_{\boldsymbol{o}} (\boldsymbol{o})\sim p_{\boldsymbol{o'}}(\boldsymbol{o'})$, $p_{\boldsymbol{x}}\sim p_{\boldsymbol{o}} (\boldsymbol{o}+\boldsymbol{c})$, and $p_{\boldsymbol{x'}} \sim p_{\boldsymbol{o'}}(\boldsymbol{o'}+\boldsymbol{c})$ where $\boldsymbol{c}=c\, \boldsymbol{1}_{p}$ is a constant vector in $R^p$, then the estimated function $f(\cdot)$ using GP with the kernel function $k_{AEN-RBF}(\boldsymbol{x}, \boldsymbol{x'})$ and the estimated function using GP with the kernel function $k_{RBF}(\boldsymbol{x}, \boldsymbol{x'})$ have the same mean squared prediction error. \\
(\textit{ii}) if $p_{\boldsymbol{o}} (\boldsymbol{o})\nsim p_{\boldsymbol{o'}}(\boldsymbol{o'})$, $p_{\boldsymbol{x}}\sim p_{\boldsymbol{o}}(\boldsymbol{o}+\boldsymbol{c})$,  $p_{\boldsymbol{x'}} \sim p_{\boldsymbol{o'}}(\boldsymbol{o'}+\boldsymbol{c})$, and $||\boldsymbol{o}-\boldsymbol{o}'||=O_p(1)$, then the estimated function $f(\cdot)$ using GP with $k_{AEN-RBF}(\boldsymbol{x}, \boldsymbol{x'})$ has a smaller mean squared prediction error than the estimated function $f(\cdot)$ using $k_{RBF}(\boldsymbol{x}, \boldsymbol{x'})$.
\\ \textbf{\textit{Proof.}}

(\textit{i}) We can re-express $\boldsymbol{x}=\boldsymbol{o}+\boldsymbol{c}$ and $\boldsymbol{x'}=\boldsymbol{o'}+\boldsymbol{c}$ without any loss of generality. Then, $\boldsymbol{x}-\boldsymbol{x'}=\boldsymbol{o}-\boldsymbol{o'}$. We can see that AEN-RBF and RBF kernels do not depend on the constant $\boldsymbol{c}$ since they only depend on distances. In this case, AEN-RBF is a symmetric kernel function and is equivalent to the RBF kernel. 

(\textit{ii}) Let $\boldsymbol{x_*}$ be a test point in $\cal{D}_X$. Let $\hat{f}_{A}$ and $\hat{f}_R$ be the estimated function using GP with AEN-RBF kernel function and RBF kernel function, respectively. We let $K_{AEN-RBF}$ and $K_{RBF}$ be the \textit{Gram matrices} associated with AEN-RBF and RBF kernel functions, respectively. For convenience, we further denote 
\begin{gather*}
K_{A}=[K_{AEN-RBF}+\sigma_{\epsilon}^2\textbf{I}_n],\\ K_{R}=[K_{RBF}+\sigma_{\epsilon}^2\textbf{I}_n], \\
k_{A_*} = k_{AEN-RBF}(\boldsymbol{x_*},\boldsymbol{x_*}), \\
k_{R_*} = k_{RBF}(\boldsymbol{x_*},\boldsymbol{x_*}), 
\end{gather*}
where $\sigma^2_{\epsilon}$ is assumed to be known and $\hat{f}_A=k_{A_*} K^{-1}_{A}\boldsymbol{y}$ and $\hat{f}_R=k_{R_*} K^{-1}_{R}\boldsymbol{y}$ are known as the least square kernel machine estimators.

Then, the mean squared prediction error difference ($MSPE$) between $\hat{f}_R$ and $\hat{f}_A$ is calculated as follows:

\small
\begin{align*}
MSPE &= E[||f(\boldsymbol{x_*})-\hat{f}_R(\boldsymbol{x_*})||^2_2-||f(\boldsymbol{x_*})-\hat{f}_A(\boldsymbol{x_*})||_2^2] \\
&=E[||f(\boldsymbol{x_*})-k_{R_*}K^{-1}_{R}\boldsymbol{y}||^2_2-||f(\boldsymbol{x_*})-k_{A_*}K^{-1}_{A}\boldsymbol{y}||^2_2]\\
&=-2\:k_{A*}K_R^{-1}k_{R*}+k_{R*}K^{-1}_R K_A K_R^{-1}k_{R*}\\
&+2\:k_{A*}K^{-1}_A k_{A*}-k_{A*}K^{-1}_A K_{A}K_{A}^{-1}k_{A*}\\
&=k_{R_*}K_{R}^{-1}K_{A}K_{R}^{-1}k_{R_*}-2k_{A_*}K_R^{-1}k_{R*}+k_{A_*}K_{A}^{-1}k_{A_*}\\
&=\{k_{R_*} (K^{-1}_R K_A K^{-1}_R)^{\frac{1}{2}} - k_{A_*}(K^{-1}_R K_A K^{-1}_R)^{-\frac{1}{2}}\}^T \\
&\times \{k_{R_*} (K^{-1}_R K_A K^{-1}_R)^{\frac{1}{2}} - k_{A_*}(K^{-1}_R K_A K^{-1}_R)^{-\frac{1}{2}}\},
\end{align*}

which is a quadratic form. 

Since $MSPE \ge 0$ and 
\begin{gather*}
k_{R_*}(K^{-1}_R K_A K^{-1}_R)^{\frac{1}{2}}-k_{A_*}(K^{-1}_R K_A K^{-1}_R)^{-\frac{1}{2}}\ne 0,
\end{gather*}
we conclude that $\hat{f}_{A}$ has a smaller mean squared prediction error than $\hat{f}_{R}$.

\section{Properties of the AEN-RBF Kernel}
\label{sec:property}
In this section, we will investigate several properties of the proposed AEN-RBF kernel, including convergence, computational complexity, and extendability to other applications.

\subsection{Convergence to the Optimum}
\begin{figure}[!t]   
 \centering
 \includegraphics[width=\columnwidth]{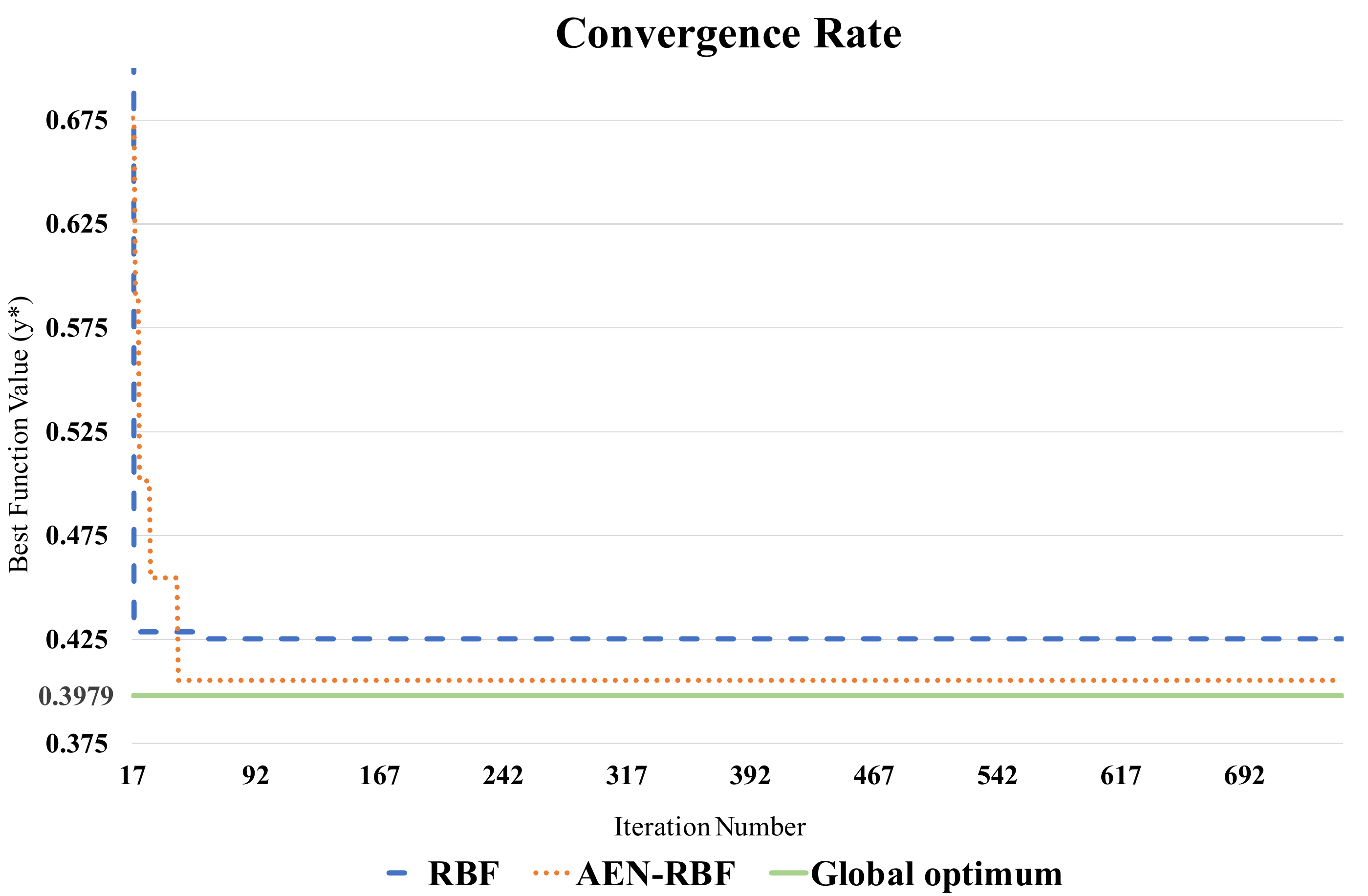}

 \caption{\textbf{Computational convergence rates of Bayesian optimization with RBF and AEN-RBF.} The convergence rates of the Bayesian optimization model when RBF and AEN-RBF kernel functions are shown in blue dashed line and orange dotted line, respectively. The convergence rate to the global minimum of the Bayesian optimization model with AEN-RBF is faster than that of the RBF.
}

 \label{fig:RBF_AEN-RBF_750}
\end{figure}

We implement one computational investigation of the convergence. The convergence rates of Bayesian optimization when RBF and AEN-RBF kernel functions are used are shown in Fig.~\ref{fig:RBF_AEN-RBF_750}. As one example, we use a 2D synthetic optimization function, Branin, to demonstrate the faster convergence to the global minimum of Bayesian optimization with AEN-RBF kernel function in comparison with Bayesian optimization with RBF kernel function. The two Bayesian optimization models are run one time for a maximum of 750 iterations where after each iteration the best function value thus far (i.e., $y^*$) is recorded. Fig.~\ref{fig:RBF_AEN-RBF_750} shows how fast Bayesian optimization approaches the global minimum when the AEN-RBF and RBF kernel functions are used. We start from the $17^{th}$ iteration for a better visualization of the convergence rates. After 750 optimization iterations, the difference between the estimated minimum and the global minimum of Branin is 0.0272 when RBF kernel function is used and 0.0071 when AEN-RBF kernel function is used. Although Bayesian optimization with AEN-RBF does not reach the global minimum after the 750 iterations, it shows a faster convergence to the global minimum as opposed to Bayesian optimization with the RBF kernel function.

\subsection{Computational Complexity}
The computational complexity of an algorithm can be measured using the number of arithmetic operations applied to a given matrix. In Bayesian optimization with Gaussian processes, the posterior mean and covariance of the Gaussian process model can be found by inverting the kernel matrix $K+\sigma_{\epsilon}^2 I_n$, where K is the \textit{Gram Matrix} corresponding to the kernel function $k$. To invert an $n \times n$ matrix, a total number of $n^3$ operations is needed (i.e., we need $\frac{n^3}{2}$ total operations for multiplication and addition operations, respectively). The computational complexity associated with $n^3$ operations is found to be $O(n^3)$. If the kernel matrix is symmetric, which is the case with most kernel matrices, the total number of operations is reduced to $\frac{n^3}{2}$ however, the computational complexity is still $O(n^3)$. In the case of an asymmetric kernel matrix, such as the AEN-RBF kernel matrix, the total number of arithmetic operations needed would be $n^3$ and thus the computational complexity would be the same as with the symmetric matrix, that is the computational complexity is $O(n^3)$.  

\begin{figure}[!t]   
 \centering
 \includegraphics[width=\columnwidth]{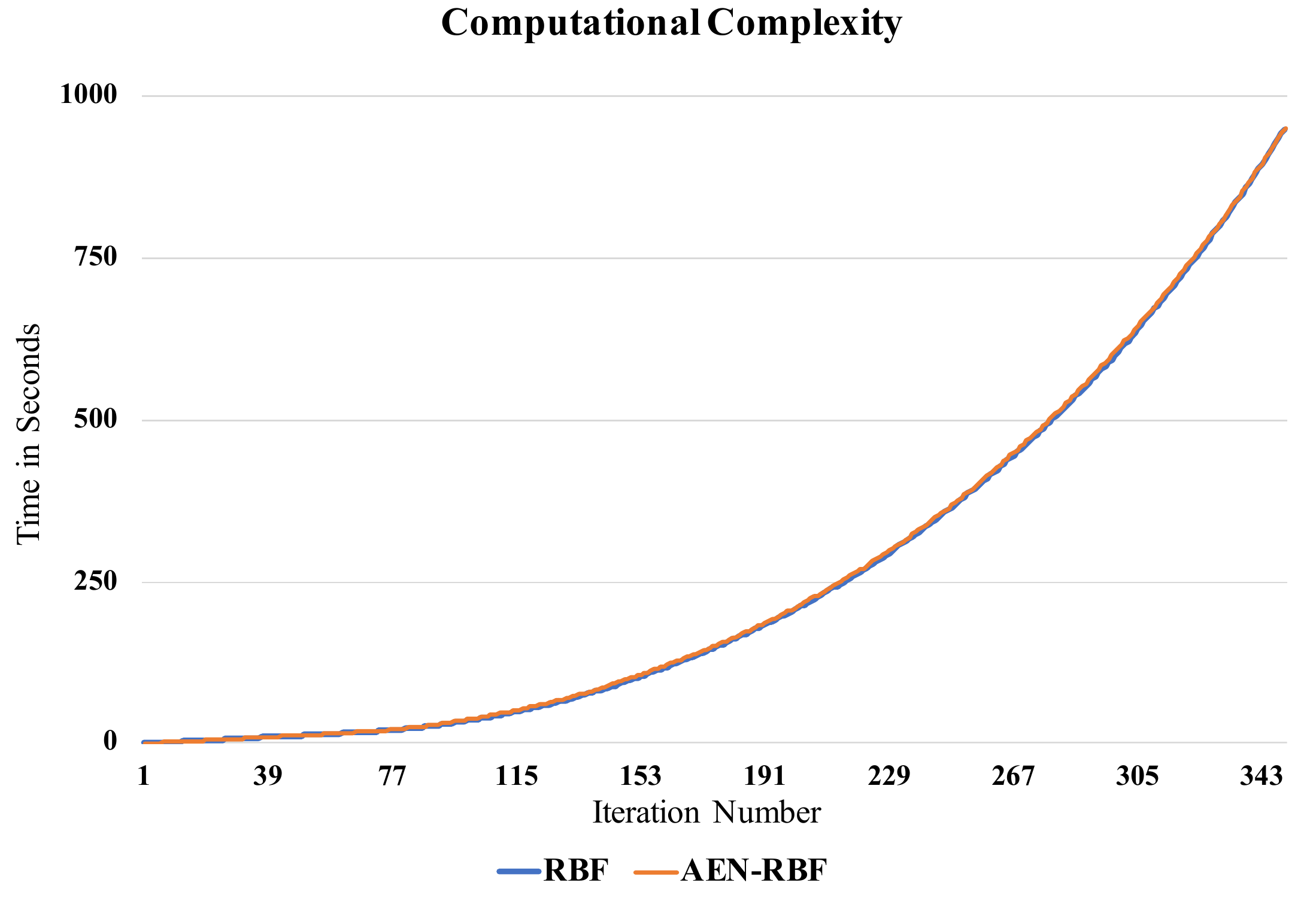}  

 \caption{\textbf{The computational time of Bayesian optimization with RBF and AEN-RBF.} The computational time of the Bayesian optimization model with RBF and AEN-RBF kernel functions are shown in blue and orange colors, respectively. The computational complexity of the Bayesian optimization model with AEN-RBF is similar to that of the RBF.
}

 \label{fig:computation_time}
\end{figure}

To show the computational performance, we also investigate the computational complexity via numerical study. We validate the theoretical computational complexities of using AEN-RBF and RBF kernel functions in Bayesian optimization by calculating the total time needed to perform 350 optimization iterations. The experiment was performed on Branin. The two Bayesian optimization models are run 10 number of times where in each run we fix the maximum number of iterations at 350. The time per iteration is recorded and averaged across the 10 runs. The average time in seconds per iteration is plotted against the number of iterations.
The computational time of Bayesian optimization with AEN-RBF kernel function in comparison to Bayesian optimization with RBF kernel function is shown in Fig.~\ref{fig:computation_time}. Fig.~\ref{fig:computation_time} shows similar computational time per iteration of Bayesian optimization with AEN-RBF and Bayesian optimization with RBF kernel functions, which is consistent with the computational complexity analysis. 

\subsection{Generalized Application of AEN-RBF in Other Domains}

The use of our proposed AEN-RBF kernel function could be generalized and further extended to other applications in machine  learning. Beside the use of kernel functions in Bayesian optimization and hyperparameter tuning of machine learning models, kernel functions could also be used as kernel tricks (e.g., in Support Vector Machine, Kernel PCA, Kernel Ridge Regression) or as activation functions in deep learning models.
\subsubsection{{Kernel Tricks}} There exists many machine learning algorithms (e.g., Support Vector Machine, Kernel PCA, Kernel Ridge Regression, etc.) where kernel tricks are used to improve the model and capture the nonlinear patterns in data. We take Support Vector Machine (SVM) as one example. When using the kernel trick~\cite{RBF_trick}, a kernel function will be used as a replacement of the dot product of two vectors in high dimensional feature spaces. The most commonly used kernel functions in SVM include the Linear, Polynomial, and the RBF kernel functions. Our proposed AEN-RBF kernel function could be generalized and used as a kernel trick in SVM to aid in capturing the nonlinear patterns and improving the flexibility and robustness of such models. The mathematical formula of the AEN-RBF kernel function used as a kernel trick in SVM is shown in Equation~\ref{eq:AEN-RBF_kernel_trick}.

\begin{equation}
\label{eq:AEN-RBF_kernel_trick}
k_{AEN-RBF-SVM}(\boldsymbol{x}, \boldsymbol{x}^{\prime}) =
\begin{cases}
  e^{-(1-\alpha) \cdot \frac{ D(\boldsymbol{x}, \boldsymbol{x}^{\prime})}{\sigma^2}} & \text{for}\: x_{ja} < x_{jb}^{\prime} \\
  e^{-\alpha \cdot \frac{D(\boldsymbol{x}, \boldsymbol{x}^{\prime})}{\sigma^2}} & \text{for}\: x_{ja} \geq x_{jb}^{\prime},
\end{cases}
\end{equation}

where $D_{E}(\boldsymbol{x}, \boldsymbol{x}')$, $D(\boldsymbol{x},\boldsymbol{x}')$, and $\alpha$ represent the squared Euclidean distance shown previously in Equation~\ref{eq:EucDis}, the weighted sum of squared Euclidean and Manhattan distances shown previously in Equation~\ref{eq:ENDis}, and the skewness parameter, respectively. The parameter $\sigma$ does not represent the variance here but rather it is a free parameter.

\subsubsection{{Activation Functions}}
Beside the use of AEN-RBF kernel function as a kernel trick in SVM, AEN-RBF could be used as an activation function in deep learning models. The mathematical formula of the  AEN-RBF activation function is shown in Equation~\ref{eq:AEN-RBF_kernel_SVM}.

\begin{equation}
\label{eq:AEN-RBF_kernel_SVM}
k_{AEN-RBF-AF}(\boldsymbol{x}, {h}_{r}) =
\begin{cases}
  e^{-(1-\alpha) \cdot \frac{ D(\boldsymbol{x}, {h}_{r})}{\sigma^2_{r}}} & \text{for}\: x_{ja} < {h}_{r} \\
  e^{-\alpha \cdot \frac{D(\boldsymbol{x}, {h}_{r})}{\sigma^2_{r}}} & \text{for}\: x_{ja} \geq {h}_{r},
\end{cases}
\end{equation}

where $D_{E}(\boldsymbol{x}, {h}_{r})$, $D(\boldsymbol{x}, {h}_{r})$, and $\alpha$ represent the squared Euclidean distance shown previously in Equation~\ref{eq:EucDis}, the weighted sum of squared Euclidean and Manhattan distances shown previously in Equation~\ref{eq:ENDis}, and the skewness parameter, respectively. Whereas, ${h}_{r}$ and ${\sigma^2_{r}}$ represent the center and the width of the hidden neuron r of the hidden layer, respectively.

\section{Numerical experiments} \label{sec:results}

Our sequential model-based optimization algorithm is defined by a Gaussian process (GP) as a surrogate model and an expected improvement (EI) as an acquisition function with L-BFGS as the optimizer. The Gaussian process model is defined by a constant as the prior mean function and our proposed AEN-RBF as the prior kernel function. We evaluate our AEN-RBF kernel function against four benchmarks, RBF, Matern 5/2, Linear, and Polynomial kernel functions. We apply Bayesian optimization on synthetic and real-world optimization functions and include the details of the setup and results of each experiment in the following subsections. When we run the Bayesian optimization, we model the Gaussian process using GPy and GPyOpt Python packages~\cite{gpy2014,gpyopt2016}.

\subsection{Optimization of Synthetic Functions}

\begin{figure}[!t]   
 \centering
 \includegraphics[width=\columnwidth]{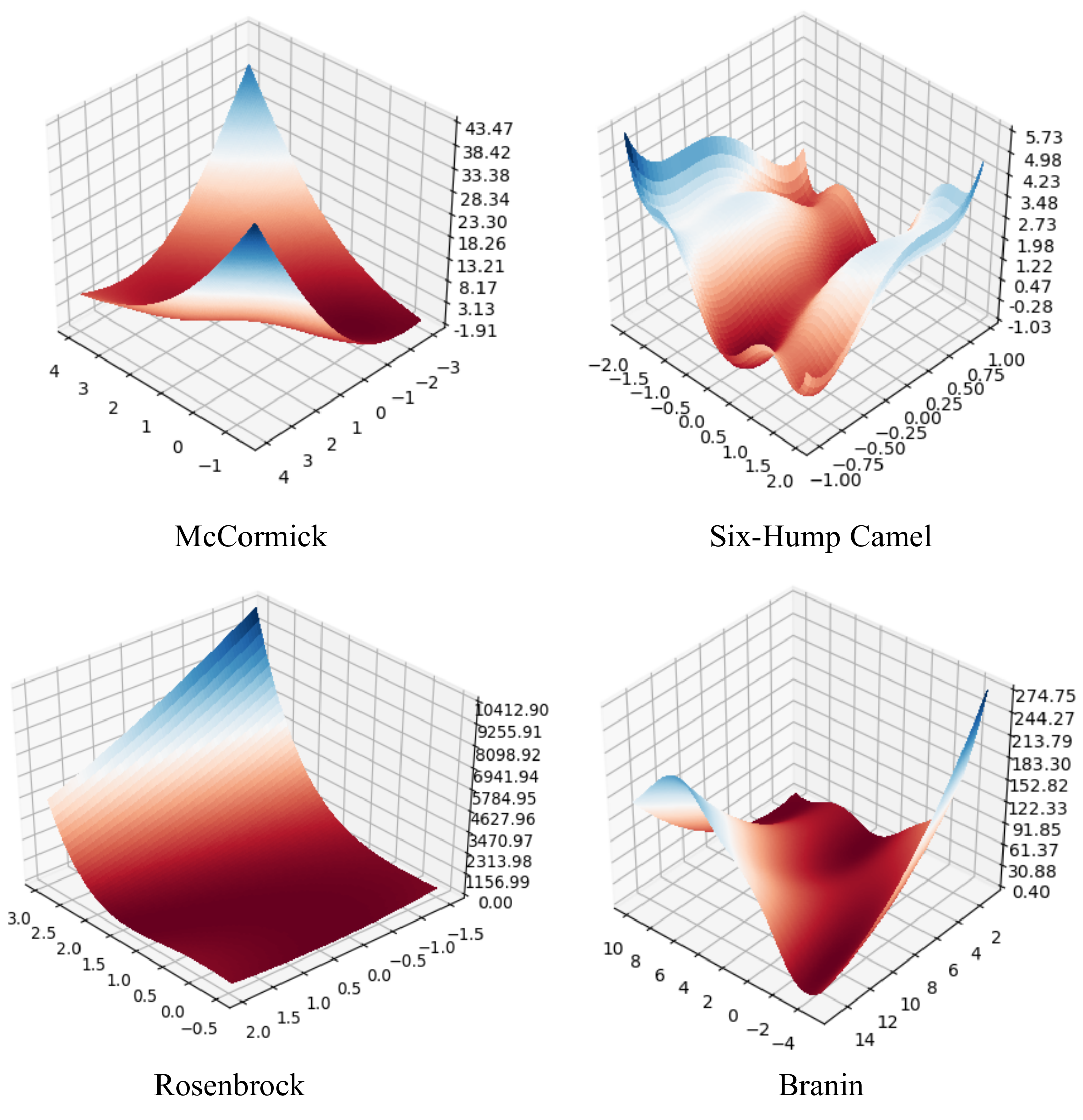}

 \caption{{Visualization of the four synthetic optimization functions.}
}
 \label{fig:Synthetics}
\end{figure}

\begin{table}[!t]
\renewcommand{\arraystretch}{1.3}
\footnotesize
\centering
\caption{Equations, global minimums and domains of four 2-dimensional benchmark functions.}
\resizebox{\columnwidth}{!}{
\begin{tabular}{|c|c|c|c|}
\hline 
\textbf{Benchmark Function} & \textbf{Equation}   &\textbf{Global Minimum}    &\textbf{Domain}      \\ 
\hline
{McCormick}    & {{{$sin(x+x') \: + \: (x-x')^2 - 1.5x + 2.5x' +1$}}} & {{{$-1.9133$}}} & $x \in (-1.5, 4)$ \\
 & & & $x' \in (-3, 4)$ \\ 
 \hline
{Six-Hump Camel}     & {$(4-2.1x^2 +\frac{x^4}{3})\:x^2 \: + \: xx' \: + \: (4x'^2 - 4)\:x'^2 $}    & {$-1.0316$} & $x \in (-2, 2)$ \\
 & & & $x' \in (-1, 1)$  \\ 
 \hline
{Rosenbrock} & {$100\:(x'-x^2)^2 \: + \: (x-1)^2$}    & {$0.0$} & $x \in (-0.5, 3)$ \\
 & & & $x' \in (-1.5, 2)$ \\ 
 \hline
{Branin}    & {$(x' - \frac{5.1}{4\pi^2}\: x^2 +\frac{5}{\pi}\: x -6)^2 \: + \: 10(1-\frac{1}{8\pi})\: cos(x) \: + \: 10$}    & {$0.3979$} & $x \in (-5, 10)$ \\
 & & & $x' \in (1, 15)$ \\ 
 \hline
\end{tabular}}
\label{tab:synthetic}
\end{table}
Bayesian optimization is used to find the global optimum of four synthetic optimization functions. More specifically, we use four two-dimensional test functions to evaluate the five kernel functions, RBF, Matern 5/2, Linear, Polynomial, and our proposed AEN-RBF. The four synthetic functions include McCormick, Six-Hump Camel, Rosenbrock, and Branin. These synthetic functions are visualized in Fig.~\ref{fig:Synthetics}.
The full description of the benchmarks along with their equations are shown in Table~\ref{tab:synthetic}.

\subsubsection{Experimental setup} To perform our experiments, we make the following design choices: (a) we fix the number of maximum optimization runs (i.e., the number of objective function evaluations) at 20, (b) we fix the number of Bayesian optimization model repetitions (i.e., the number of times the sequential model-based optimization algorithm is run) at 30, (c) we fix the initial points of the optimization model for all kernels for fair comparison, 
and (d) we perform three experiment scenarios, primarily, Bayesian optimization without any induced outliers, Bayesian optimization with 5\% induced outliers, and Bayesian optimization with 10\% induced outliers.

\textbf{Metrics.} 
We use the root mean squared error (RMSE) as a metric to evaluate AEN-RBF against the three benchmark kernel functions. Equation~\ref{eq:rmse} represents the mathematical formula of RMSE, where $y_{r}^*$ is the best estimated function value of the rth repetition and R represents the total number of Bayesian optimization model repetitions. A lower RMSE value implies that the Bayesian optimization model predicts an estimated global minimum closer to the true global minimum of the synthetic optimization function, and hence produces a reliable predictive distribution over the objective function. Therefore, a lower RMSE value is better.
\begin{equation}
\label{eq:rmse}
RMSE = \sqrt{\sum_{r=1}^{R} (y_{r}^{*} - y_{global\:min})^2}
\end{equation}

\subsubsection{Input Space Domain in the Presence of Outliers} We validate the robustness of our proposed kernel function, AEN-RBF, using two outlier scenarios. We perform Bayesian optimization with 5\% and 10\% induced outliers. In the case of 5\% induced outliers, both $x$ and $x'$ domain intervals are widened by 2.5\% on both sides of each interval whereas in the 10\% induced outliers, both $x$ and $x'$ domain intervals are widened by 5\% on both sides of each interval.

Table~\ref{tab:OD_synthetics} shows the input domains (i.e., interval domains of $x$ and $x'$) of each synthetic function in the 5\% and 10\% induced outlier scenarios.

\begin{table}[!t]
\renewcommand{\arraystretch}{1.3}
\footnotesize
\centering
\caption{The domains of $x$ and $x'$ in the two considered outlier cases, 5\% and 10\% induced outliers, respectively.}
\resizebox{\columnwidth}{!}{
\begin{tabular}{|c|c|c|}
\hline
\textbf{Benchmark Functions} & \textbf{Domain with 5\% induced outliers} & \textbf{Domain with 10\% induced outliers} \\ 
\hline
\textbf{McCormick} & {$x \in (\text{\textminus 1.59}, 4.2), x' \in (\text{\textminus 3.17}, 4.2)$}  &   {$x \in (\text{\textminus 1.8}, 4.31), x' \in (\text{\textminus 3.39}, 4.39)$} \\
\hline
\textbf{Six-Hump Camel}   & {$x \in (\text{\textminus 2.1}, 2.11), x' \in (\text{\textminus 1.05}, 1.06)$} & {$x \in (\text{\textminus 2.22}, 2.22), x' \in (\text{\textminus 1.11}, 1.11)$} \\ 
\hline
\textbf{Rosenbrock}  & {$x \in (\text{\textminus 0.59}, 3.1), x' \in (\text{\textminus 1.59}, 2.11)$}  & {$x \in (\text{\textminus 0.69}, 3.2), x' \in (\text{\textminus 1.69}, 2.2)$}  \\ 
\hline
\textbf{Branin}   & {$x \in (5.39, 10.4), x' \in (0.63, 15.37)$} &   {$x \in (5.67, 11), x' \in (0.44, 16)$} \\ 
\hline
\end{tabular}}
\label{tab:OD_synthetics}
\end{table}

\subsubsection{Results}

We evaluate our kernel function AEN-RBF against the three benchmarks on three experiment scenarios. Table~\ref{tab:RMSE_Synthetics} shows the RMSE values of each kernel function tested on the four synthetic optimization functions without any induced outliers. From Table~\ref{tab:RMSE_Synthetics}, we see that the AEN-RBF kernel function outperforms RBF, Matern,  , and Polynomial kernel functions in all the four test functions (i.e., McCormick, Six-Hump Camel, Rosenbrock, and Branin). Matern refers to Matern 5/2 kernel function. We also report the RMSE values of the two remaining experiments with the 5\% and 10\% induced outliers in Table~\ref{tab:RMSE_Synthetics_outliers}.
\begin{table}[!t]
\renewcommand{\arraystretch}{1.3}
\footnotesize
\centering
\caption{Results for The Root Mean Squared Error (RMSE) values of each kernel function evaluated on four synthetic optimization functions. The minimum RMSE value obtained in each experiment is shown in bold.}
\resizebox{\columnwidth}{!}{
\begin{tabular}{|c|c|c|c|c|}
\hline
\textbf{Kernel Function} & \textbf{McCormick} & \textbf{Six-Hump Camel} & \textbf{Rosenbrock} & \textbf{Branin} \\ \hline
\textbf{RBF}             & 0.0140             & 0.0115                  & 0.5524              & 0.2996          \\ \hline
\textbf{Matern}          & 0.0501             & 0.0115                  & 0.4579              & 0.4154          \\ \hline
\textbf{Linear}          & 0.0655             & 0.0173                  & 0.4968              & 0.4436          \\ \hline
\textbf{Polynomial}      & 0.0117             & 0.0156                  & 0.6381              & 0.6845          \\ \hline
\textbf{AEN-RBF}         & \textbf{0.0100}    & \textbf{0.0083}         & \textbf{0.3526}     & \textbf{0.2124} \\ \hline
\end{tabular}}
\label{tab:RMSE_Synthetics}
\end{table}

The reduction in RMSE values is due to the combination of Manhattan and Euclidean distances as well as the use of the skewness parameter $\alpha$. The combination brings together the advantages of Manhattan and Euclidean distances which can balance the exploitation and exploration better. Meanwhile the skewness parameter adds flexibility to the model thus yields reliable robust Bayesian optimization.

\begin{table}[!t]
\renewcommand{\arraystretch}{1.3}
\footnotesize
\centering
\caption{Results for The Root Mean Squared Error (RMSE) values of each kernel function evaluated on four synthetic optimization functions. 5\% and 10\% represent five percent and ten percent induced outliers, respectively. The minimum RMSE value obtained in each experiment is shown in bold.}
\resizebox{\columnwidth}{!}{
\begin{tabular}{|c|c|c|c|c|c|c|c|c|}
\hline
 \textbf{Kernel Functions} & \multicolumn{2}{c|}{\textbf {McCormick}} & \multicolumn{2}{c|}{\textbf {Six-Hump Camel}} & \multicolumn{2}{c|}{\textbf {Rosenbrock}} & \multicolumn{2}{c|}{\textbf {Branin}} \\ \cline{2-9} 
                            & \textbf {5\%}          & \textbf {10\%}          & \textbf {5\%}       & \textbf {10\%}             & \textbf {5\%}         & \textbf {10\%}           & \textbf {5\%}        & \textbf {10\%}        \\ \hline
\textbf{RBF}                        & 0.0183            & 0.0159             & 0.0185          & 0.0605           & 0.5508           & 0.6712           & 1.0614          & 0.7166           \\ \hline
\textbf {Matern}                         & 0.1599            & 0.0213             & 0.0225          & 0.0622           & 0.6128           & 0.7225           & 0.4837          & 0.8930           \\ \hline
\textbf {Linear}                         & 0.0164              & 0.0157             & 0.0261          & 0.0345         &0.6492            & 0.6728           & 0.8157          & 0.5658           \\ \hline
\textbf {Polynomial}                     & 0.0139              & 0.0244             & 0.0351          & 0.0339        & 0.6148            & 0.7067           & 0.6272          & 0.6120           \\ \hline
\textbf {AEN-RBF}                         & \textbf {0.0113}            & \textbf {0.0110}             & \textbf {0.0124}          & \textbf {0.0166}           & \textbf {0.4216}           & \textbf {0.5597}           & \textbf {0.3684}          & \textbf {0.5049}           \\ \hline
\end{tabular}}
\label{tab:RMSE_Synthetics_outliers}
\end{table}

\subsection{Hyperparameter Tuning of Deep Learning Models}
We use Bayesian optimization to tune hyperparameters of deep learning models. The hyperparameters of such models are not tuned automatically when deep learning models are trained, on the contrary, they have to be configured manually and optimized using hyperparameter tuning algorithms. In this paper, we use Bayesian optimization with Gaussian processes to tune those hyperparameters. We explain the tuned deep learning model in details as well as the experimental setup in the next subsections. We report the accuracy of the deep learning model when each of the four kernel functions is used. Since our goal is to obtain a higher accuracy, we minimize the negative of the objective function that is, we minimize the negative of deep learning models accuracy.

\begin{figure}[!t]   
 \centering
 \includegraphics[width=\columnwidth]{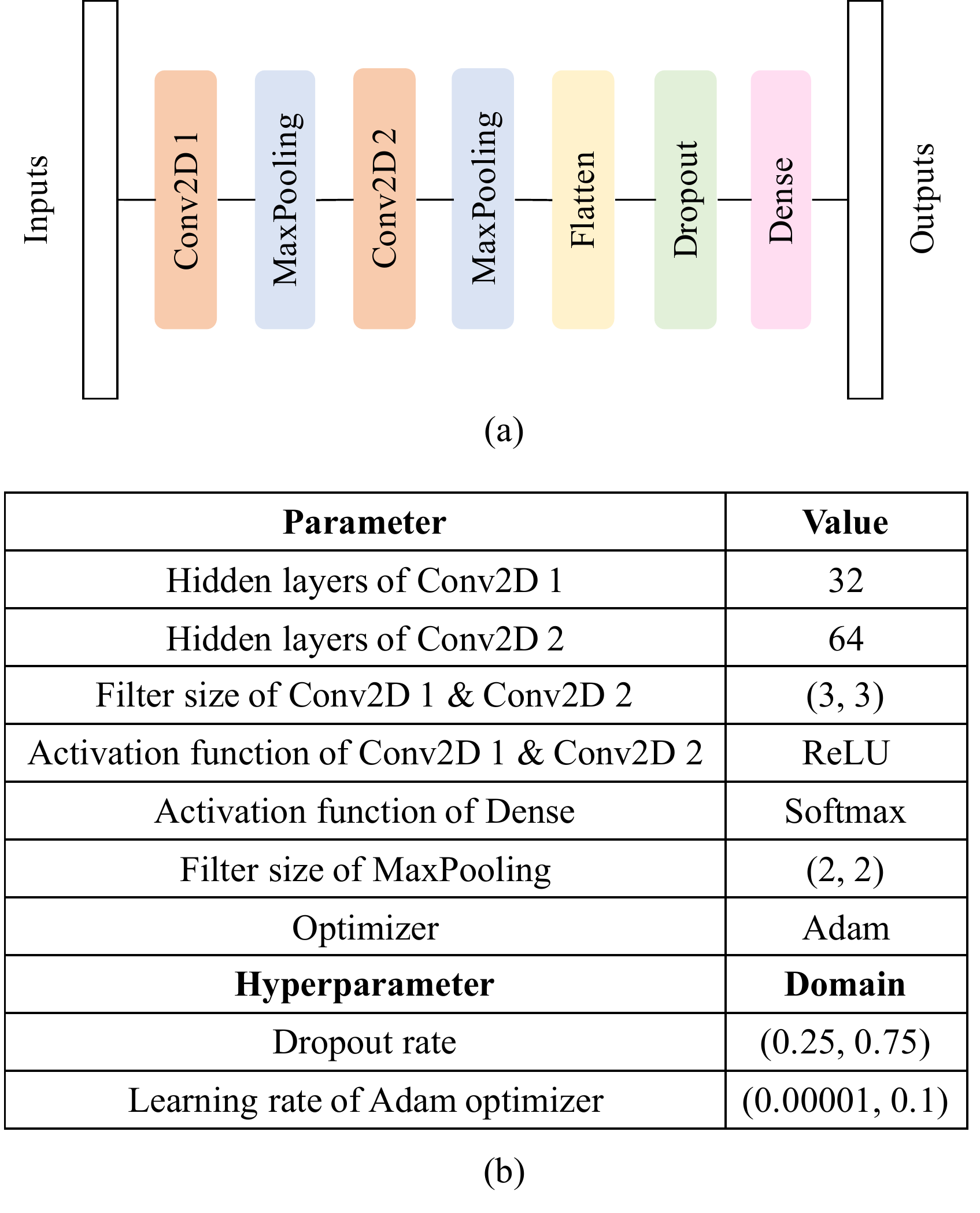}

 \caption{\textbf{Feed Forward Neural Network (FFNN).} (a) Represents the FFNN architecture and (b) represents the parameters and hyperparameters of the FFNN
}

 \label{fig:FFNN}
\end{figure}

\textbf{Feed Forward Neural Network.}
We use a feed-forward neural network (FFNN) to evaluate AEN-RBF kernel function against RBF, Matern, Linear, and Polynomial benchmark kernels. The FFNN architecture is shown in Fig.~\ref{fig:FFNN}. The FFNN is used to classify input images to their respective classes. The experimental setup along with the image dataset details are included in the following subsections.

\subsubsection{Experimental setup} We use Bayesian optimization with GPs to optimize the FFNN. The FFNN deep learning model is used for classification purposes, more specifically, we use images as inputs and classify them into their respective classes. We use convolutional layers followed by maximum pooling layers as shown in subfigure (a) in Fig.~\ref{fig:FFNN}. We fix the FFNN parameters for all the associated FFNN models of the four kernels. FFNN parameters are included in subfigure (b) in Fig.~\ref{fig:FFNN}. We use ReLU activation functions after each convolutional layer and a dropout layer before the final dense layer. We use Adam optimizer and a softmax activation function for the final dense layer. We optimize two hyperparameters, the dropout rate of the dropout layer and the learning rate of Adam optimizer. We include the domain of each hyperparameter in subfigure (b) in Fig.~\ref{fig:FFNN}. 
For the Bayesian optimization model, we make the following design choices: (a) we train the FFNN deep learning model for 10 epochs, (b) we run the Bayesian optimization model 20 times, (c) we use 10 maximum function evaluations per Bayesian optimization run, and (d) we fix the batch size at 128. We perform two experiment scenarios to evaluate our AEN-RBF kernel function against the three benchmarks. More specifically, we apply Bayesian optimization to optimize the FFNN without and with induced outliers.

\textbf{Dataset.} We use the MNIST dataset to demonstrate the effectiveness of our kernel function in optimizing the FFNN in the two experiment scenarios. The MNIST dataset consists of a training set of 60,000 images and a test set of 10,000 images. The images are gray scale and have a size of 28*28 pixels.

\subsubsection{Input Space Domain in the Presence of Outliers}
We validate the robustness of our proposed kernel function, AEN-RBF, using one outlier scenario besides the case without any induced outliers. We perform Bayesian optimization to the Feed-Forward Neural Network with an enlarged learning rate domain (i.e., learning rate $\in (0, 1)$) which is considered very wide in comparison with the standard learning rate domain.

\subsubsection{Results} We evaluate our AEN-RBF kernel function against RBF, Matern, Linear, and Polynomial benchmark kernels in two experiment scenarios. We use the average classification accuracy of the FFNN as a metric for evaluating AEN-RBF kernel against the three benchmark kernels. We report the average accuracy for each kernel function in Table~\ref{tab:MNIST_FFNN}. From Table~\ref{tab:MNIST_FFNN}, we see that our AEN-RBF kernel function has the highest average accuracy in hyperparameter tuning of the FFNN in both experiments (i.e., without and with induced outliers), and thus AEN-RBF outperforms the three benchmark kernels.

\begin{table}[!t]
\renewcommand{\arraystretch}{1.3}
\footnotesize
\centering
\caption{Results for the average classification accuracy of each benchmark kernel function evaluated on the feed-forward neural network with the MNIST dataset. The maximum average accuracy in each experiment is shown in bold.}
\resizebox{\columnwidth}{!}{ 
\begin{tabular}{|c|c|c|} 
\hline
\textbf{Kernel Function}    & \textbf{Without Induced Outliers}         & \textbf{With Induced Outliers}    \\ 
\hline
\textbf{RBF}    & 98.63\%    & 89.91\%   \\ 
\hline
\textbf{Matern}  & 98.68\%   & 97.94\%   \\ 
\hline
\textbf{Linear}  & 98.93\%   & 98.20\%    \\ 
\hline
\textbf{Polynomial} & 98.71\%  & 97.91\%   \\ 
\hline
\textbf{AEN-RBF} & \textbf{98.95\%}  & \textbf{98.38\%} \\ 
\hline

\end{tabular}}
\label{tab:MNIST_FFNN}
\end{table}

\section{Case Study: Image Defect Detection in Advanced Manufacturing}
\label{sec:case_study}

Early defect detection in advanced manufacturing systems is crucial for maintaining safe operations of machines and improving the quality of the manufactured products. Steel manufacturing in particular is one of the most important industrial processes with steel being the building block of many industries including but not limited to construction, machinery, automotive, oil, and gas industries. Surface defect detection and prediction in steel help improve and upgrade the quality of the final product. While deep learning models have been widely used in the detection, localization, and prediction of steel defects, the uniqueness of different manufacturing datasets ensure that hyperparameters of deep learning models need to be adjusted before training and testing, hence the strong need for hyperparameter tuning through Bayesian optimization which takes into consideration the different characteristics of steel defects.

\subsection{Experimental setup}
We use Bayesian optimization with Gaussian processes to optimize the feed-forward neural network (FFNN) deep learning model shown in subfigure (a) in Fig~\ref{fig:FFNN}. We perform two experiment scenarios, without and with induced outliers, to evaluate AEN-RBF kernel function against the three benchmark kernels. We seek to optimize two hyperparameters, the learning rate of Adam optimizer and the dropout rate as shown in subfigure (b) in Fig~\ref{fig:FFNN}. In the second experiment (i.e., FFNN with induced outliers), we use a very large learning rate domain while the dropout rate domain remains the same.

For the Bayesian optimization model setup: (a) we use 20 epochs to train the FFNN deep learning model, (b) we use 10 Bayesian maximum iterations (i.e., function evaluations), (c) we run the Bayesian optimization model 20 times, and for fair comparison, (d) we fix the initial points of the Bayesian optimization model for all kernel functions.

\begin{figure}[!t]   
 \centering
 \includegraphics[width=\columnwidth]{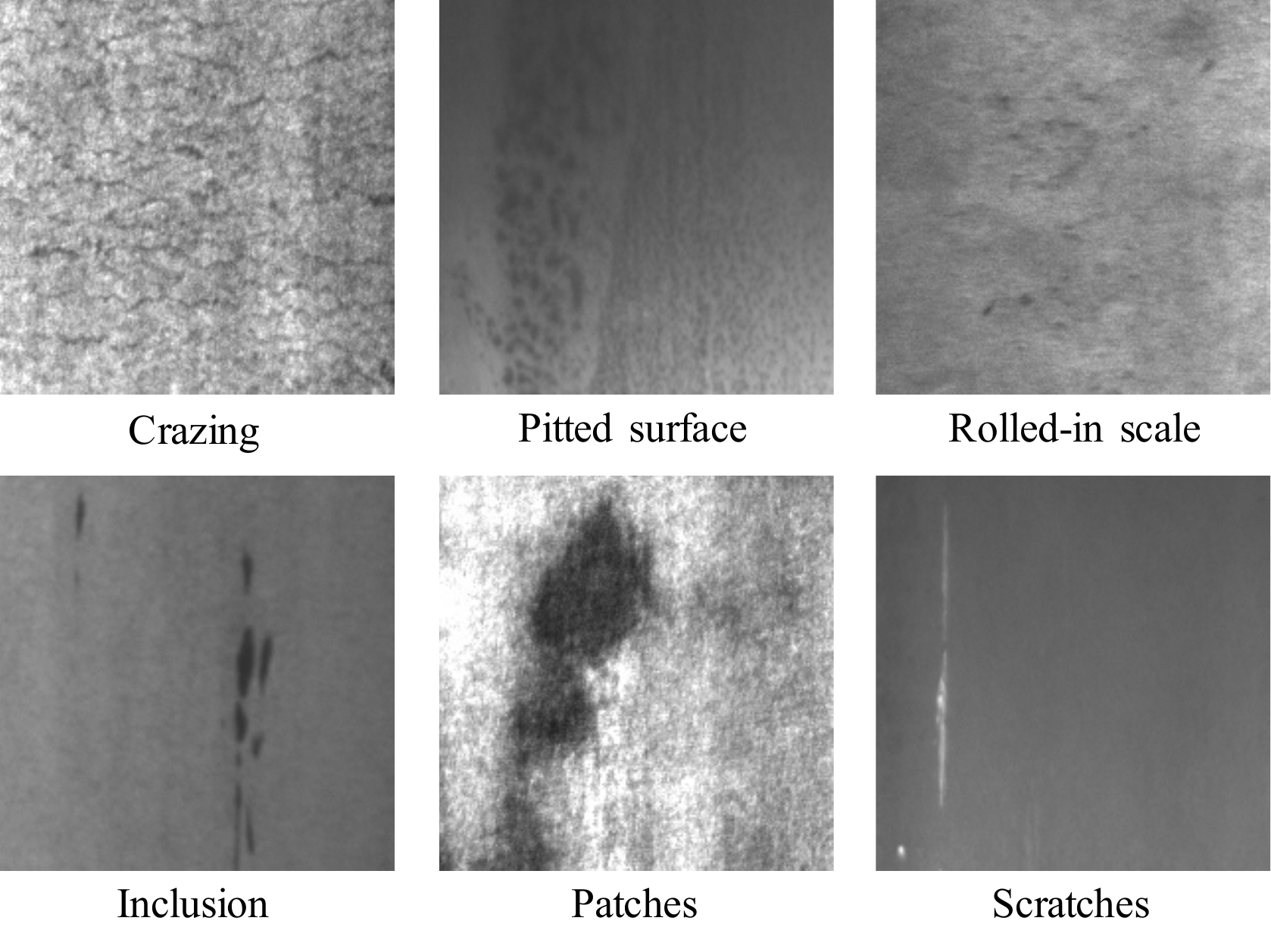}

 \caption{{The six types of steel surface defects of the NEU dataset.}
}
 \label{fig:Steel_defect}
\end{figure}

\textbf{Dataset.} The dataset consists of 1800 images, each of which is a gray scale with a size of 200*200 pixels. The number of images of the dataset is divided equally on six types of steel surface defects (i.e., 300 images per defect type). The six types of steel surface defects are rolled-in scale, patches, crazing, pitted surface, inclusion, and scratches. A sample of the six types of steel surface defects is shown in Fig.~\ref{fig:Steel_defect}.
\subsection{Results}

\begin{table}[!t]
\renewcommand{\arraystretch}{1.3}
\footnotesize
\centering
\caption{Results of the average classification accuracy of each benchmark kernel function evaluated on the feed-forward neural network with the steel defect dataset. The maximum average accuracy in each experiment is shown in bold.}
\resizebox{\columnwidth}{!}{ 
\begin{tabular}{|c|c|c|} 
\hline
\textbf{Kernel Function} & \textbf{Without Induced Outliers}           & \textbf{With Induced Outliers}      \\ 
\hline
\textbf{RBF}  & 74.74\%    & 79.53\%       \\ 
\hline
\textbf{Matern}   & 75.56\%   & 79.39\%     \\ 
\hline
\textbf{Linear}    & 67.65\%   &  74.54\%    \\ 
\hline
\textbf{Polynomial}   & 69.15\%  &  76.36\%   \\ 
\hline
\textbf{AEN-RBF}  & \textbf{81.28\%}  & \textbf{88.18\%}    \\ 
\hline
\end{tabular}}
\label{tab:Steel_FFNN}
\end{table}
We evaluate our AEN-RBF kernel function against the four benchmark kernels, RBF, Matern, Linear, and Polynomial. We use the average classification accuracy as a metric to evaluate the five benchmarks. The goal is to classify the six types of steel surface defects correctly and achieve a high accuracy. Table~\ref{tab:Steel_FFNN} represents the average accuracy of each kernel function when optimizing the FFNN on the steel defects dataset. From Table~\ref{tab:Steel_FFNN}, our AEN-RBF achieves the highest average accuracy in the two experiment scenarios. We conclude that the AEN-RBF kernel function outperforms the RBF, Matern, Linear, and Polynomial kernel functions.

\subsection{Stability of the hyperparameter tuning}
\begin{figure}[!t]   
 \centering
 \includegraphics[width=\columnwidth]{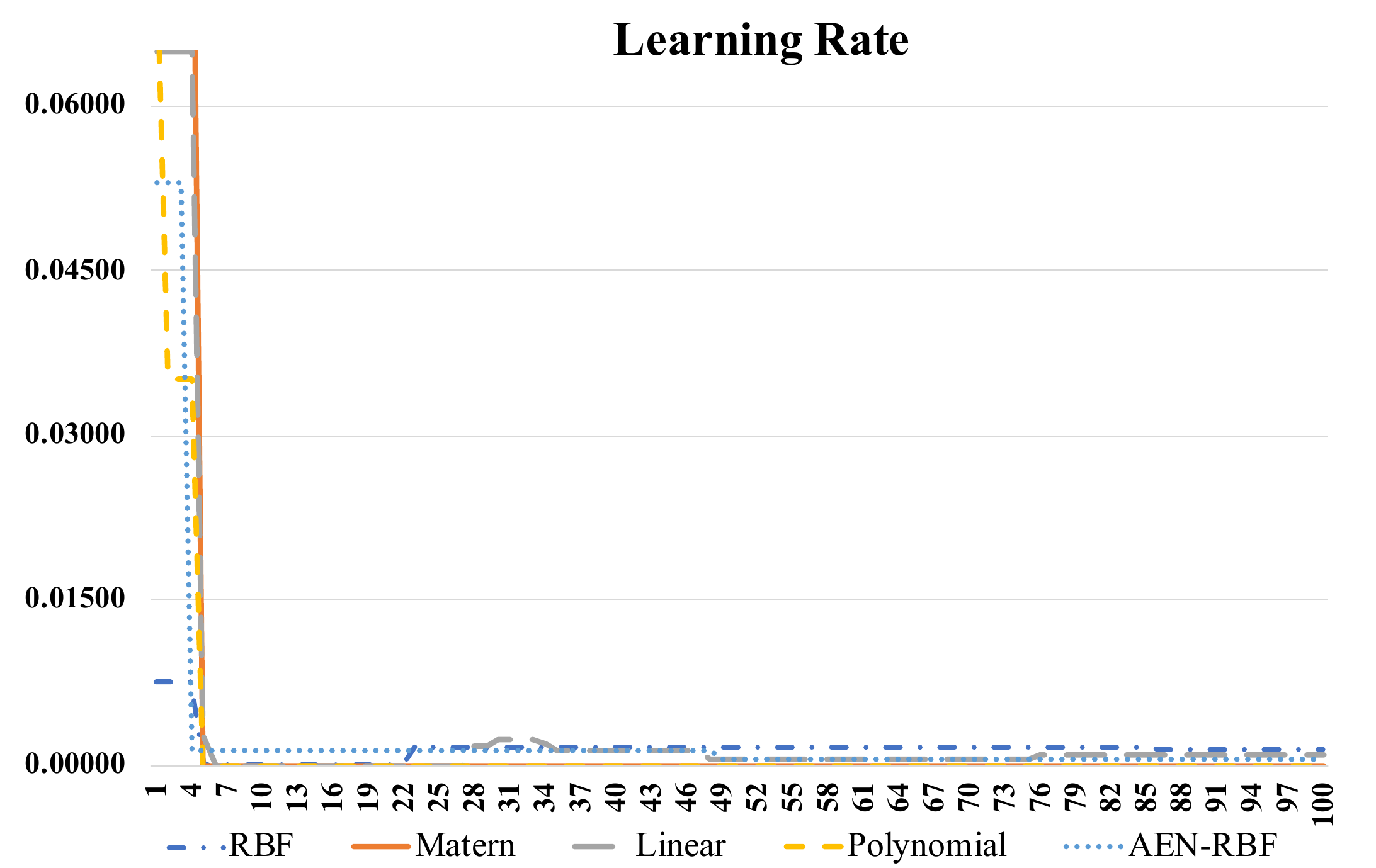}  

 \caption {The convergence rates of optimizing the learning rate hyperparameter of the FFNN with Bayesian optimization using the five benchmark kernel functions.
}

 \label{fig:LR}
\end{figure}

\begin{figure}[!t]   
 \centering
 \includegraphics[width=\columnwidth]{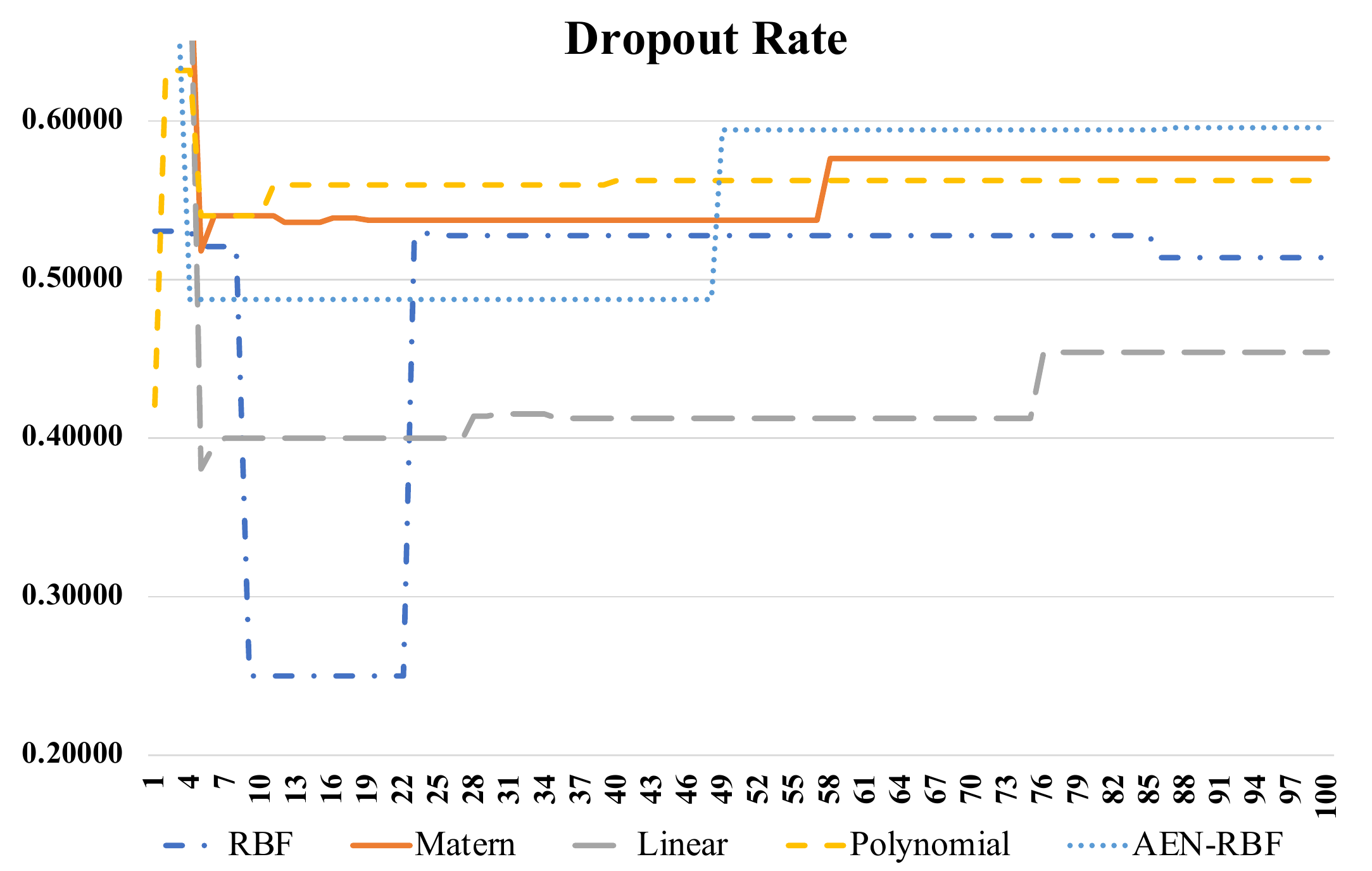}  

 \caption{The convergence rates of optimizing the dropout rate hyperparameter of the FFNN with Bayesian optimization using the five benchmark kernel functions.
}

 \label{fig:DR}
\end{figure}

We also evaluate the stability and convergence of estimating the hyperparameters (learning rate and dropout rate)  of the FFNN when the five kernel functions, namely, RBF, Matern, Linear, Polynomial, and AEN-RBF are used. We run the Bayesian optimization model one time for a maximum of 100 optimization iterations and record the optimal values of the hyperparameters in each iteration. Fig.~\ref{fig:LR} and Fig.~\ref{fig:DR} show the stability of estimating the FFNN hyperparameters after 100 iterations. We can find that both the learning rate and dropout rate converge to a stable value after some iterations. The convergence speed of learning rate is higher than the convergence speed of dropout rate. 


\section{Conclusion}
\label{sec:conclusions}
Bayesian optimization is a probabilistic approach used to optimize black-box functions. It is critical for complex engineering systems where the objective function is expensive to evaluate, has an unknown analytical form, and is highly nonlinear and non-convex. When optimizing unknown black-box functions with Gaussian processes, the choice of the kernel function has a direct impact on the performance of Bayesian optimization. Existing kernels are sensitive and susceptible to outliers, which may cause the surrogate model to be inaccurate, thus yielding unreliable function approximations and system optimization. 
We proposed AEN-RBF, a novel asymmetric robust kernel function for Bayesian optimization with Gaussian processes. We proved its validity as a kernel function and showed its effectiveness in reducing the predictive mean squared error theoretically. We analyzed the convergence to global optimum of the AEN-RBF kernel function as well as its computational efficiency in comparison with the RBF kernel function. 

The proposed AEN-RBF kernel function can improve the robustness and flexibility of Bayesian optimization. We generalized the use of AEN-RBF kernel function to other machine learning domains such as the use of AEN-RBF as a kernel trick and as an activation function in feed forward neural networks. 
Extensive evaluations on four synthetic functions and a deep learning model revealed that, compared to RBF, Matern, Linear, and Polynomial kernels, our AEN-RBF kernel function achieves lower RMSE values and higher classification accuracy, respectively. As a case study of hyperparameter tuning in the manufacturing domain, we applied Bayesian optimization with Gaussian processes to tune a feed-forward neural network on a steel surface defect dataset. We evaluated our kernel function against the four benchmarks and showed that AEN-RBF outperforms existing approaches.

\ifCLASSOPTIONcaptionsoff
  \newpage
\fi



%



\bibliographystyle{IEEEtran}
\bibliography{main}

%

\begin{IEEEbiography}[{\includegraphics[width=1in,height=1.25in,clip,keepaspectratio]{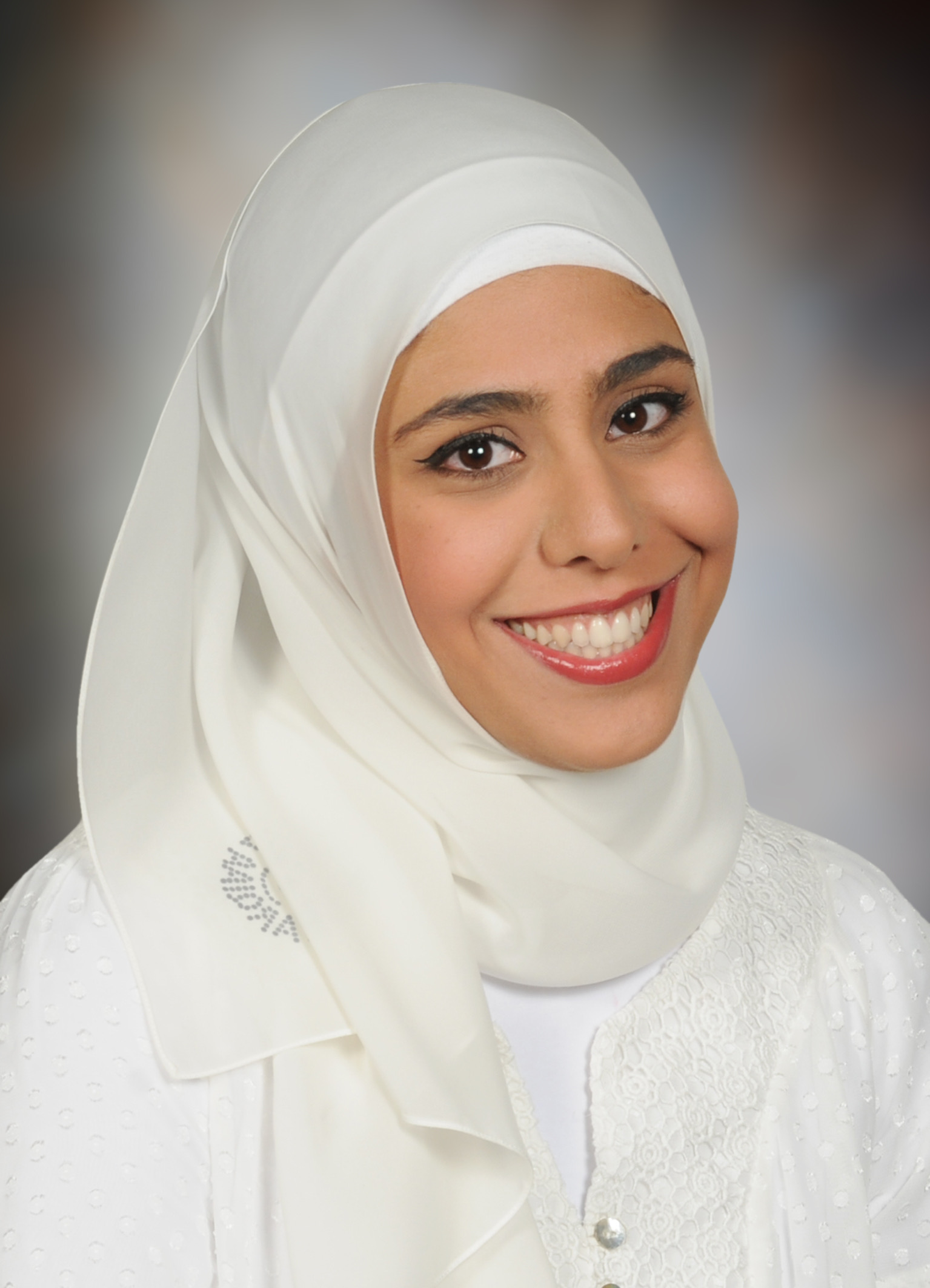}}]{Areej AlBahar}
received her B.S. degree in Industrial and Management Systems Engineering from Kuwait University, Kuwait, in 2012. She received her M.S. degree in Industrial and Systems Engineering from the University of Florida, Gainesville, USA, in 2016. She is currently working toward her Ph.D. degree at the Grado Department of Industrial and Systems Engineering, Virginia Tech, Blacksburg, USA.

Her research interests include machine learning and advanced statistics for monitoring, optimization and quality improvement of manufacturing processes. She is a recipient of NAMRC/MSEC NSF Student Award. 
\end{IEEEbiography}

\begin{IEEEbiography}[{\includegraphics[width=1in,height=1.25in,clip,keepaspectratio]{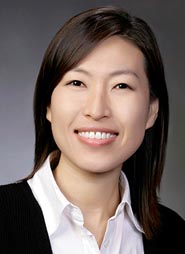}}]{Inyoung Kim} received the B.S. degree in Mathematics from the Jeju National University, Jeju, Korea, in 1994, the M.S. in Applied Statistics from the Yonsei University, Seoul, Korea, in 1996, Ph.D. in Statistics from the Texas A\&M University, College Station, USA, in 2002 and Postdoc in Biostatistics from the Yale University, New Haven, USA, in 2015. 

Currently, she is an associate professor at the Department of Statistics, Virginia Tech, Blacksburg, USA. Her research interests are focused on developing statistical machine learning and semi/nonparametric statistical methods and theory for high dimensional analysis. Both Frequentist and Bayesian methods have been developed. She is a recipient of Best Paper Award in the year of \textit{Biometrics} 2016. Dr. Kim is a member of ASA, IBS, and ISBA.

\end{IEEEbiography}

\begin{IEEEbiography}[{\includegraphics[width=1in,height=1.25in,clip,keepaspectratio]{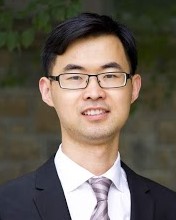}}]{Xiaowei Yue}
(S’15-M’18-SM'21) received the B.S. degree in mechanical engineering from the Beijing Institute of Technology, Beijing, China, in 2011, the M.S. in power engineering and engineering thermophysics from the Tsinghua University, Beijing, China, in 2013, the M.S. in Statistics, Ph.D. in Industrial Engineering with minor Machine Learning  from the Georgia Institute of Technology, Atlanta, USA, in 2016 and 2018. 

Currently, he is an assistant professor at the Grado Department of Industrial and Systems Engineering, Virginia Tech, Blacksburg, USA. His research interests are focused on machine learning for advanced manufacturing. He is a recipient of FTC Early Career Award, eight best paper awards (e.g., \textit{IEEE Transactions on Automation Science and Engineering} Best Paper Award), and two best dissertation awards. He was a Mary and Joseph Natrella Scholar from ASA. Yue also won several teaching awards (e.g., Large Class Teaching Award) from Center for Excellence in Teaching and Learning at Virginia Tech. He is a DoD MEEP Faculty Fellow. He also serves as an associate editor for the \textit{IISE Transactions} and the \textit{Journal of Intelligent Manufacturing}.  
    
Dr. Yue is a senior member of ASQ, IEEE, and IISE, and a member of ASME and SME. 

\end{IEEEbiography}







\end{document}